\documentclass[lettersize,journal]{IEEEtran}
\usepackage{amsmath,amsfonts}
\usepackage{algorithmic}
\usepackage{algorithm}
\usepackage{cite}
\usepackage{array}
\usepackage[caption=false,font=normalsize,labelfont=sf,textfont=sf]{subfig}
\usepackage{textcomp}
\usepackage{color,bm}
\usepackage{booktabs}
\usepackage{threeparttable}
\usepackage{stfloats}
\usepackage{url}
\usepackage{multirow}
\usepackage{verbatim}
\usepackage{graphicx}
\def\BibTeX{{\rm B\kern-.05em{\sc i\kern-.025em b}\kern-.08em
		T\kern-.1667em\lower.7ex\hbox{E}\kern-.125emX}}
\usepackage{balance}

\usepackage{amsthm}
\usepackage{amssymb}
\theoremstyle{definition}

\newtheorem{owndef}{\bf Definition}
\newtheorem{ownprop}{\bf Proposition}

\begin{document}
	\title{Slender Object Scene Segmentation in Remote Sensing Image Based on Learnable Morphological Skeleton with Segment Anything Model}
	\author{Jun Xie, Wenxiao Li, Faqiang Wang, Liqiang Zhang, Zhengyang Hou, Jun Liu
    \thanks{Faqiang Wang, Liqiang Zhang and Jun Liu were supported by the National Natural Science Foundation of China (No.42293272), Faqiang Wang and Jun Liu were also supported by  the National Natural Science Foundation of China (No.12371527) and the Beijing Natural Science Foundation (No. 1232011). (Jun Xie, Wenxiao Li contributed equally to this work.) (Corresponding author: Jun Liu.)
    
    Jun Xie, Wenxiao Li, Faqiang Wang, Jun Liu are with the School of Mathematical Sciences, Beijing Normal University, Beijing, 100875, China (e-mail: xiej@mail.bnu.edu.cn, 202431130055@mail.bnu.edu.cn, fqwang@bnu.edu.cn, jliu@bnu.edu.cn).

    Liqiang Zhang, Zhengyang Hou are with the State Key Laboratory of Remote Sensing Science, Faculty of Geographical Science, Beijing Normal University, Beijing, 100875, China (e-mail: zhanglq@bnu.edu.cn, houzy@mail.bnu.edu.cn).}}

	\maketitle
	\begin{abstract}
 Morphological methods play a crucial role in remote sensing image processing, due to their ability to capture and preserve small structural details. However, most of the existing deep learning models for semantic segmentation are based on the encoder-decoder architecture including U-net and Segment Anything Model (SAM), where the downsampling process tends to discard fine details. In this paper, we propose a new approach that integrates learnable morphological skeleton prior into deep neural networks using the variational method. To address the difficulty in backpropagation in neural networks caused by the non-differentiability presented in classical morphological operations, we provide a smooth representation of the morphological skeleton and design a variational segmentation model integrating morphological skeleton prior by employing operator splitting and dual methods. Then, we integrate this model into the network architecture of SAM, which is achieved by adding a token to mask decoder and modifying the final sigmoid layer, ensuring the final segmentation results preserve the skeleton structure as much as possible. Experimental results on remote sensing datasets, including buildings, roads and water, demonstrate that our method outperforms the original SAM on slender object segmentation and exhibits better generalization capability.
	\end{abstract}
	
	
	\section{Introduction}
	Semantic segmentation is a fundamental problem in remote sensing image analysis and other fields, aiming to separate objects of interest from the background at the pixel level. This technique has been widely utilized in various practical applications, including urban planning, damage detection, land cover classification, and resource management. 
	
	At present, the mainstream image segmentation methods can be broadly classified into two categories: model-based methods and data-driven methods. The former typically regards the image segmentation as an optimization problem of energy minimization, such as the Potts model \cite{potts1952some} and the Snakes model \cite{kass1988snakes}.
	These energy functions generally consist of two primary components: a fidelity term that ensures the accuracy of segmentation and a regularization term that preserves essential spatial features within the segmentation result. However, the design of traditional variational models requires substantial expertise in related fields and often suffers from poor generalization. Data-driven methods can overcome these limitations by leveraging deep learning techniques. Some neural networks were proposed in last few years like Fully Convolutional Network (FCN)\cite{long2015fully}, U-net\cite{ronneberger2015u} and Vision-Transformer (ViT)\cite{dosovitskiy2020image}, which significantly improved the flexibility and adaptability of image segmentation.
    


    
	Unlike natural images, segmentation in remote sensing images requires preserving local slender structures and overall topological properties, such as connectivity, holes, and branches, which makes this task particularly challenging. The morphological skeleton \cite{maragos1986morphological} is a simplified representation of the object’s structure that can accurately describe these topological features. Several studies have incorporated skeletons as prior knowledge in segmentation models. For instance, Hu et al. \cite{hu2009road} utilized skeletons as the initial values for the Snakes model to guide road extraction. Additionally, Gaetano et al. \cite{gaetano2011morphological} proposed a segmentation method based on the watershed transform \cite{beucher1979use}, where skeletons were used as markers. Skeleton prior has also been integrated into data-driven models. Shit et al.\cite{shit2021cldice} improved the skeleton extraction algorithm and designed a loss function that combines traditional segmentation loss with skeleton loss. Similarly, \cite{xu2021cp} proposed an connectivity-preserving loss based on skeleton for the detection of line-shaped objects. Unlike above methods, Liu et al. \cite{liu2018roadnet} designed a CNN to predict  road surface, edges, and center lines, and correlated these results to enhance road extraction. However, all these methods lack reliable mathematical interpretability, and whether the subnetworks or loss functions can learn the desired features depends entirely on empirical evidence.
	
	In recent years, foundational models like GPT \cite{brown2020language} and CLIP \cite{radford2021learning} have transformed the field of natural language processing. Recently, Meta AI introduced a visual foundational model called Segment Anything Model (SAM)\cite{kirillov2023segment}. After pre-training on the SA-1B dataset, SAM can segment any type of object using prompts inputs such as points, boxes, masks, or text. It demonstrates strong zero-shot generalization performance across multiple remote sensing datasets\cite{ren2024segment}. Currently, two main approaches exist for adapting SAM to downstream tasks: the first is to design one-shot or few-shot methods to enhance SAM's performance by learning from a limited number of examples in new datasets \cite{zhang2023personalize, osco2023segment}. The second approach involves adding adaptors or prompters\cite{chen2024rsprompter, chen2023sam} to SAM's network architecture, tailoring outputs for specific tasks, but this lacks a rigorous mathematical foundation.

    Most segmentation networks utilize an encoder-decoder architecture to extract high-dimensional features. However, the upsampling and downsampling processes in this design often lead to the loss of fine details, limiting the network's ability to segment small targets or structures. To address this issue, we integrate morphological skeleton prior into the neural network to enhance segmentation accuracy and preserve overall geometric structure. There are primarily three strategies to achieve this. The first strategy is to post-process the neural network outputs using traditional models. For instance, Zhang et al.\cite{zhu2020building} utilized Conditional Random Field (CRF) to extract spatial neighbourhood information of buildings, which was fused with the network's output to ensure clear segmentation boundaries. Yet, this approach does not correct inherent errors in the neural network's output. The second strategy focuses on constructing a loss function that incorporates spatial priors to guide the model's learning process. For example, Ma et al. \cite{ma2024sam} introduced a object consistency loss and a boundary preservation loss to fine-tune SAM. However, these models may struggle when input data significantly deviates from the training data, as they cannot use learned priors knowledge during prediction.
    The third approach unrolls the variational segmentation algorithm to design network layers, integrating prior constraints through regularization and penalty terms into the model. This technique combines the advantages of both model-based and data-driven methods, enabling the network to automatically minimize the energy functional during forward propagation. Monga et al. \cite{monga2021algorithm} described this technique in deep learning, and numerous studies have explored its potential \cite{zheng2015conditional, jia2021regularized, meng2023interpretable, liu2022deep}. Apart from unrolling methods, \cite{li2024boundary} proposed a network called BEDSN, which uses sub-networks to learn boundary priors and applies them in decoding process. Nonetheless, these methods \cite{zao2023topology, li2023semantic, li2024boundary, ma2024swint} often lack the same level of mathematical interpretability provided by variational models.

    In this study, we aim to develop a mathematically interpretable segmentation model that preserves morphological skeleton priors. Unlike existing methods that typically incorporate skeleton and other structural information within the loss function, such as the cl-dice in \cite{shit2021cldice}, we construct an energy functional that maintains the skeleton prior as a regularization term. This functional is integrated into the Soft Threshold Dynamic (STD) segmentation model \cite{liu2022deep}. By employing operator splitting and Fenchel-Legendre duality, we transform the model into three sub-problems, for which we provide both the solution form and an iterative solving algorithm. Furthermore, we unroll the variational model into a novel network module, which we refer to as the Morphological Skeleton Prior (MorSP) module. This module can be seamlessly integrated into any neural network framework, endowing the network with the ability to preserve learnable skeleton priors during the decoding process. In this paper, we demonstrate the effectiveness of our approach by incorporating the MorSP module to SAM, showcasing improvements in segmentation accuracy and detail preservation for building, road and water extraction of remote sensing images.
	
	The main contributions of this paper are as follows:
	\begin{itemize}
		\item We provide a smooth approximation of morphological operators, offering a differentiable representation of the morphological skeleton and its variation. This allows the integration of non-differentiable morphological operations into the backpropagation process within neural networks.
		\item We design a variational segmentation model based on smooth morphological skeletons and Threshold Dynamics, which can be efficiently solved using the operator splitting method. This facilitates the incorporation of learnable morphological skeleton priors into any neural network architecture.
		\item We propose a novel approach to integrate the morphological skeleton prior into the Segment Anything Model (SAM). This fusion method is mathematically interpretable and can automatically learn and preserve the skeleton prior during both training and prediction across various datasets. Additionally, we demonstrate its advantages in building, road and water extraction tasks through ablation experiments.
	\end{itemize}
	
	The paper is organized as follows: Section II reviews related works, including variational segmentation models, classical morphological operators, and the Segment Anything Model. Section III presents the proposed segmentation model that preserves morphological skeleton priors and introduces the MorSP module, detailing its integration into SAM. Section IV evaluates the performance of MorSP Module on remote sensing images. Section V presents numerical experiment results and analysis. Finally, we summarize and discuss this work.
	
	\section{Related Works}
	\subsection{Variational Segmentation Methods}
	Mathematically, a remote sensing image is represented as a map $I: \Omega\rightarrow\mathbb{R}^d$ on the pixel domain $\Omega$, where $d=1$ means a gray-scale image, and $d>1$ means multispectral image. Therefore, segmentation tasks can be seen to divid the pixel domain $\Omega$ into many distinct parts: $\Omega=\cup_{i=1}^N\Omega_i,\Omega_i\cap\Omega_j=\emptyset, i\ne j$, $N$ is the total classes. Subregion $\Omega_i$ can be represented as its relax indicative function $u_i:\Omega\rightarrow[0,1]$ and the segmentation condition can be written as a simplex
	\begin{equation*}
		\mathbb{U} = \{ \mathbf{u} = (u_1,\dots,u_N)\in[0,1]^N: \sum\limits_{i=1}^N u_i(x)=1,\forall x\in \Omega \}.
	\end{equation*}
	
	The Potts model\cite{potts1952some} is one of the most classic variational segmentation models and it solves
	\begin{equation}
		\label{potts}
		\min\limits_{\Omega_i}-\sum\limits_{i=1}^N\int_{\Omega_i}o_i(x)u_i(x)dx + \lambda\sum\limits_{i=1}^N|\partial \Omega_i|,
	\end{equation}
	where $o_i(x)$ denotes the similarity of pixel $x$ to the $i$-th class. The second term is a spatial regularization term, in which $\lambda>0$ is a parameter and $ |\partial \Omega_i| $ denotes the length of boundary of $\Omega_i$. Usually, Total Variation (TV)\cite{rudin1992nonlinear} is applied to represent the length term in segmentation models\cite{chan2001active, bresson2007fast}.
	
	Due to the non-smooth and high calculation cost of TV, \textcolor{black}{a threshold dynamics (TD) method that expresses boundaries as differences in values within neighborhoods has been proposed. The idea of TD originally stemmed from the spatially-based Markov Random Field \cite{Geman1984Stochastic}, and was later applied to spatial regularization in fuzzy, K-means, and Gaussian mixture model types based image segmentation \cite{PHAM2001Spatial, Wang2009, Liu2011, Liu2013}. The advantage of the threshold dynamics method is that it can be efficiently solved by the MBO splitting scheme \cite{merriman1992diffusion} and applied to PDE-based image segmentation \cite{esedog2006threshold, wang2017efficient}. The TD energy to approximate the boundary length can be written as:
	}
	\begin{equation}
		\label{TDterm}
		|\partial\Omega_i|\approx\sqrt{\frac{\pi}{\sigma}}\sum\limits_{i=1}^N\int_{\Omega}u_i(x)(f\ast (1-u_{i}))(x)dx,
	\end{equation}
	where $\ast$ denotes convolution and $f$ is a kernel such as Gaussian function $f(x)=\frac{1}{2\pi \sigma^2}e^{-\frac{||x||^2}{2\sigma^2}}$.
	
	Liu et al \cite{liu2022deep} proposed a Soft Threshold Dynamics (STD) model based the above TD method \eqref{TDterm} using twice Fenchel-Legendre transformation:
	\begin{equation}
		\label{STD}
		\hat{\mathbf{u}} = \arg \min\limits_{\mathbf{u}\in\mathbb{U}} \{ \langle-\mathbf{o},\mathbf{u}\rangle+\gamma \langle \mathbf{u}, \ln \mathbf{u}\rangle + \mathcal{R}(\mathbf{u}) \}.
	\end{equation}
	where $ \mathcal{R}(\mathbf{u}) = \lambda\langle \mathbf{u},f\ast(1-\mathbf{u})\rangle$. Since $\mathcal{R}$ is concave when $f$ is positive definite, they offer the following iterative formula to solve this problem by using the difference of convex (DC) algorithm:
	\begin{equation*}
		\mathbf{u}^{t+1} = \arg \min\limits_{\mathbf{u}\in\mathbb{U}}\{\langle-\mathbf{o},\mathbf{u}\rangle + \gamma\langle \mathbf{u}, \ln \mathbf{u}\rangle + \mathcal{R}(\mathbf{u}^t)+\langle \mathbf{p}^t , \mathbf{u} - \mathbf{u}^t \rangle \},
	\end{equation*}
	where $\mathbf{p}^t = \lambda f * (1-2\mathbf{u}^t) \in \partial \mathcal{R}(\mathbf{u}^t)$. This method not only has lower computational complexity and higher stability than TV regularization, but also allows the addition of other variational priors like star-shape and volume priors. Its stability and computational efficiency will greatly improve the efficiency of segmenting large-scale remote sensing images. Besides, the STD provides the relationship between the commonly used classification function softmax in neural networks and the variational method for image segmentation.
	
	\subsection{Classic Morphological Image Processing}
	The operators of mathematical morphology are more effective for extracting structural information than normal convolution operations, as they are directly related to the shapes of objects. Operators of gray-scale morphology can be defined using mathematical formalism\cite{haralick1987image}.
	
	\begin{owndef}[gray-scale dilation and erosion\cite{gonzales1987digital}]
		\label{def:g-sop}
	Supposed that $I: \Omega \rightarrow \mathbb{R}$ is an image, structuring element $\mathbb{B}$ is a subset of $\Omega$ and $(0,0)\in\mathbb{B}$ . The dilation of $I$ by $\mathbb{B}$ is denoted by $I\oplus \mathbb{B}$ and is defined as
		\begin{equation*}
			(I\oplus \mathbb{B})(x,y) := \max\limits_{(x^{'},y^{'})\in \mathbb{B}}\{ I(x+x^{'}, y+y^{'}) \}.
		\end{equation*} 
		Similarly, the erosion of $I$ by $\mathbb{B}$ is denoted by $I\ominus \mathbb{B}$ and is defined as
		\begin{equation*}
			(I\ominus \mathbb{B})(x,y) := \min\limits_{(x^{'},y^{'})\in \mathbb{B}}\{ I(x+x^{'}, y+y^{'}) \}.
		\end{equation*}
	\end{owndef}
	
\begin{figure}[!t]
		\centering
		\includegraphics[width=3.5in]{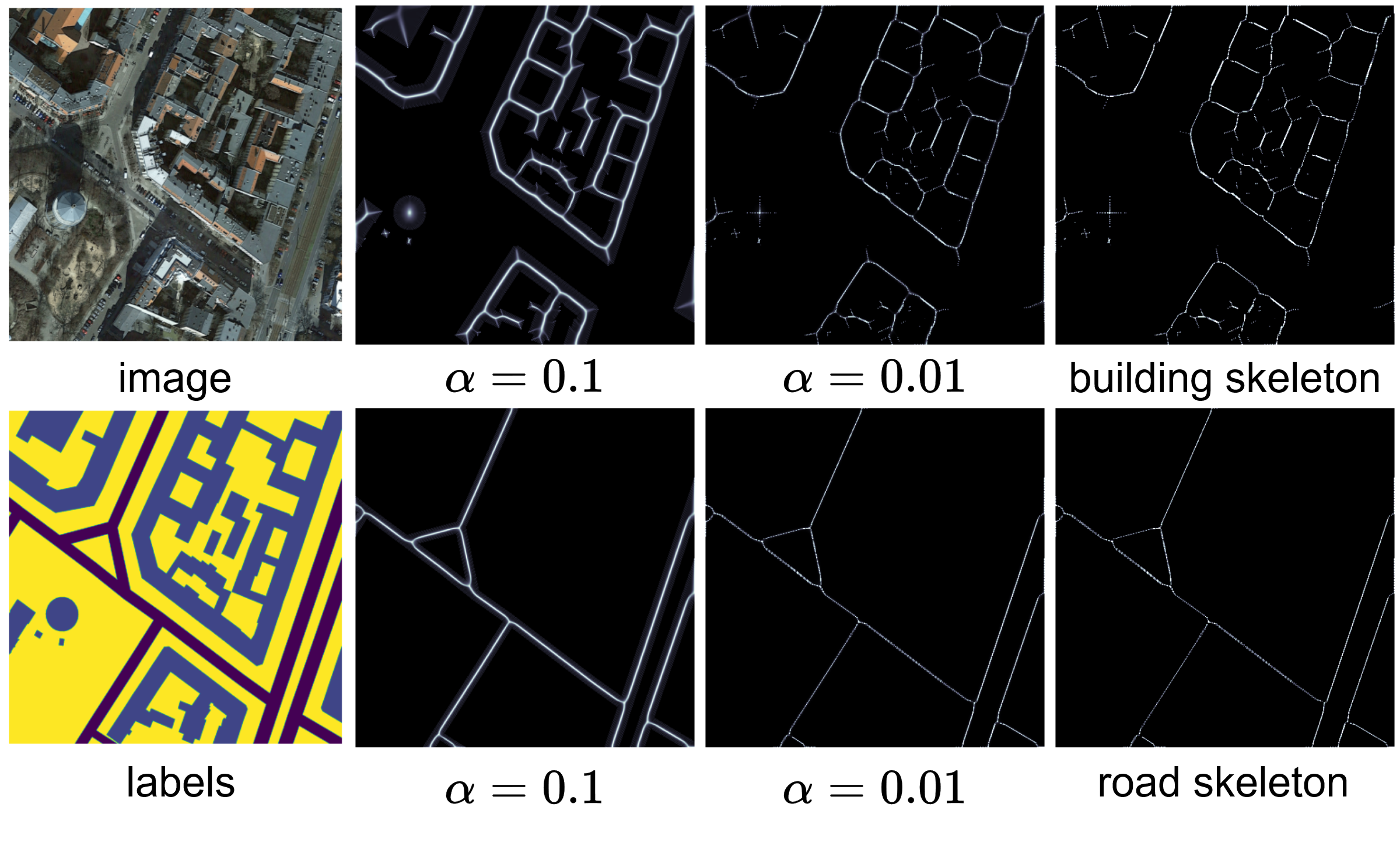}
		\caption{Illustration of skeleton extraction: from left to right are original image and labels, smooth skeletons, classic skeletons.}
		\label{fig:skelex}
	\end{figure}

	While dilation and erosion operators individually may not play a significant role, their combination can lead to advanced and practical algorithms. The skeleton of gray-scale image can be represented by dilation and erosion operation:
    \begin{equation}
        \label{skel}
        \mathcal{S}(I) = \sum\limits_{j=0}^J \mathcal{S}_j(I),
    \end{equation}
    where 
    \[
        \mathcal{S}_j(I) = (I\ominus ^j\mathbb{B}) -((I\ominus^j \mathbb{B})\ominus \mathbb{B}) \oplus \mathbb{B} ,
    \]
    $J$ is the times of operator composition, $\mathbb{B}$ is a structuring element and $ (I\ominus^j \mathbb{B})$ means $I$ is eroded $j$ times with $\mathbb{B}$ continuously. The morphological skeleton method is often used in the task of segmenting mesh or elongated structures. The rightmost column in Figure \ref{fig:skelex} shows the skeleton extraction results of buildings, roads in the remote sensing images by above method. However, the operators defined in morphology inherently contain non-differentiable min and max operators, and designing them directly into network architectures would result in the absence of gradients for backpropagation. Therefore, we will address this issue in the later section.
 
	\subsection{Segment Anything Model and Fine-tuning Strategies}
	SAM\cite{kirillov2023segment} is an interactive foundation model designed for segmentation. Trained on millions of natural images, it exhibits strong zero-shot generalization capabilities. The model consists of three main components: an image encoder, a prompt encoder and a mask decoder. The image encoder is a Vision Transformer (ViT) \cite{dosovitskiy2020image} pre-trained using Masked Autoencoders (MAE) \cite{he2022masked}, which maps input images into a high-dimensional embedding space, denoted as $\Phi_{\theta_1}: \mathbb{R}^{c\times h\times w}\rightarrow\mathbb{R}^{c_1\times h_1\times w_1}$. The prompt encoder converts two sets of prompts (sparse: points, boxes; dense: masks) to 256-dimensional embeddings, denoted as $ \Phi_{\theta_2}: \mathbb{R}^k \times \mathbb{R}^{h\times w}\rightarrow \mathbb{R}^{c_1 \times k} \times \mathbb{R}^{c_1\times h_1\times w_1  }  $. The mask decoder uses a two-way transformer including cross attention and mlp layers to fuse image feature and prompt feature, denoted as  $\Phi_{\theta_3}:\mathbb{R}^{c_1\times h_1\times w_1}\times\mathbb{R}^{c_1 \times k }\rightarrow\underbrace{\mathbb{R}^{h\times w}\times\cdots\times\mathbb{R}^{h\times w}}_n$, where n denotes the number of output. Then we can write SAM as
	\begin{equation}
        \label{eq:sam}
		\begin{cases}
			F_I = \Phi_{\theta_1}(I),\ T_p, T_m = \Phi_{\theta_2}(p, m), \\
			o = \Phi_{\theta_3}(F_I + T_m, T_p),\\
			u^{\ast} = \mathcal{H}(o),
		\end{cases}
	\end{equation}
	where $m, p$ are dense and sparse prompt respectively, $F_I, T_m, T_p$ indicate image feature, dense embedding and sparse embedding, and $o$ is the segmentation feature extracted by SAM. $\mathcal{H}$ is a decoding operator that is heaviside step function here:
	\begin{equation}\label{eq:Heaviside}
		\mathcal{H}(o)(x)=\begin{cases}
			0, \quad o(x)\le 0, \\
			1, \quad o(x)>0.
		\end{cases}
	\end{equation}
	Thus, the segmentation process of SAM can be divided into three sub-problems: $\Phi_{\theta_1}, \Phi_{\theta_2}$ are encoding processes, $\Phi_{\theta_3}$ is the features fusion process, and $\mathcal{H}$ is the final decoding process.
	
	Pretrained visual foundational models like SAM excel at learning and extracting image features. Consequently, many studies have focused on leveraging SAM's strengths and adapting it to downstream tasks. Yan et al. \cite{yan2023ringmo} proposed a multimodal remote sensing segmentation model based on SAM, which can segment both optical and synthetic aperture radar (SAR) images. Chen et al.\cite{chen2024rsprompter} developed a method to automatically segment remote sensing images by adding a prompter to generate prompts suitable for SAM.
	
	
	Low-Rank Adaptation (LoRA)\cite{hu2021lora} is a fine-tuning strategy for large foundational models that involves freezing pretrained weights and introducing learnable low-rank decomposition matrices into each layer of transformer blocks. It offers the advantage of fewer learnable parameters while incurring no additional inference latency.
	For a given token $T_{in}\in\mathbb{R}^{N\times d}$ and a pretrained weight matrix $W\in\mathbb{R}^{d\times k}$, LoRA applies low-rank decomposition to update $W$, modifying the forward process to
	\begin{equation}
		T_{out} = T_{in}(W+\Delta W) = T_{in}(W + AB),
	\end{equation}
	where $ T_{out} \in \mathbb{R}^{N\times k}$ is the output token and $A\in\mathbb{R}^{d\times r}, B\in\mathbb{R}^{r\times k}$ are two learnable matrices with a low rank $r\ll \min(d,k)$. There have been several studies using LoRA to fine-tune SAM. For example, SAMed \cite{zhang2023customized}, a specialized version of SAM tailored for medical imaging, fine-tuned its image encoder using the LoRA strategy.
	
	In this study, we mainly focus on the decoding process where the morphological skeleton prior will be integrated, and the original Heaviside step function is equivalent to the minimization problem:
	\begin{equation*}
		\mathcal{H}(o) = \arg\min\limits_{u\in[0,1]}\{ \langle -o, u\rangle \}.
	\end{equation*}
	This equivalent optimization problem will be utilized in our subsequent method. 
	
	\section{The Proposed Method}
	\subsection{Smooth Morphological Skeleton}
	In order to integrate the morphological skeleton prior into our variational framework and facilitate gradient calculation, we give a smooth approximation form of morphological operator. Below $u:\Omega\rightarrow[0,1]$ stands for single-channel gray-scale image.
	
	Take a given closed rectangle or closed disk structuring element $ \mathbb{B}\ni (0,0) $, we note that gray-scale dilation and erosion operators are 
	\begin{equation*}
		\mathcal{D}(u)(x) = \max\limits_{z\in \mathbb{B}}u(x+z), \quad \mathcal{E}(u)(x) = \min\limits_{z\in \mathbb{B}}u(x+z) .
	\end{equation*}
	In order to get a smooth skeleton for gradient backpropagation in SAM, we can smooth the above operators by log-sum-exp function.
	\begin{ownprop}[smooth dilation]
        \label{th:SM}
		For $\alpha>0$, define smooth functional:
		\begin{equation}
				\mathcal{D}^{\alpha}(u)(x):= \alpha \ln\int_{z\in \mathbb{B}}e^{\frac{u(x+z)}{\alpha}} dz,
		\end{equation}
		where $\alpha>0$. If $u$ is continuous in $x+\mathbb{B}$, we have
		\begin{equation*}
			\mathcal{D}(u) = \lim\limits_{\alpha\rightarrow0^{+}} \mathcal{D}^{\alpha}(u).
		\end{equation*}
	\end{ownprop}
    \begin{IEEEproof}
        The detailed proof is shown in Appendix A.
    \end{IEEEproof}

    It is not difficulty to prove that $\mathcal{D}^{\alpha}(u)$ is convex with respect to $u$, thus we have its twice Fenchel-Legendre transformation $(\mathcal{D}^{\alpha})^{\ast\ast} = \mathcal{D}^{\alpha}$. For a fixed point $x\in \Omega$, we define the inner product on $\mathbb{B}$:
    \begin{equation*}
        \langle v, u\rangle_\mathbb{B} := \int_{\mathbb{B}}v(x+z)u(x+z) dz.
    \end{equation*}
    
    \begin{ownprop}[Smooth dilation operator]
        \label{th:SD}
        The twice Fenchel-Legendre transformation of $\mathcal{D}^{\alpha}$ is 
        \begin{equation*}
            (\mathcal{D}^{\alpha})^{\ast\ast}(u)(x) = \max\limits_{k\in\mathbb{K}(x)} \{ \langle k, u \rangle_\mathbb{B} - \alpha\langle k, \ln k\rangle_\mathbb{B} \}, \ \forall x\in\Omega,
        \end{equation*}
        where $\mathbb{K}(x) = \{k:\Omega\rightarrow[0,1],\ \int_\mathbb{B} k(x+z)dz=1 \}$. Furthermore, 
        \begin{equation*}
        \begin{split}
        \mathcal{D}^{\alpha}(u)(x) =& \int_{\Omega} (\mathcal{K}_{\mathcal{D}}(u)(x,y)u(x+y)\\&- \alpha \mathcal{K}_{\mathcal{D}}(u)(x,y)\ln\mathcal{K}_{\mathcal{D}}(u)(x,y))dy,
        \end{split}
        \end{equation*}
        where 
        \[ \]
        \[ \mathcal{K}_{\mathcal{D}}(u)(x,y) = \begin{cases}
        \frac{e^{\frac{u(x+y)}{\alpha}}}{\int_\mathbb{B} e^{\frac{u(x+z)}{\alpha}} dz}, \quad &y \in \mathbb{B}, \\
        0, &y \in \Omega \backslash \mathbb{B},
        \end{cases}
        \]
        is smooth dilation kernel.
    \end{ownprop}
    \begin{IEEEproof}
        The detailed proof is shown in Appendix B.
    \end{IEEEproof}
    
    Compared with the non-differential gray-scale dilation operator in Definition \ref{def:g-sop}, the above dual representation of dilation operator is smooth, which can prevent singularities in the backpropagation of the network.
    
    Noticing that $\mathcal{E}(u) = -\mathcal{D}(-u)$, we denote $\mathcal{E}^{\alpha}(u) = -\mathcal{D}^{\alpha}(-u)$, we have analogous dual smooth representation:
    \begin{equation*}
    \begin{split}
    \mathcal{E}^{\alpha}(u)(x) =& \int_{\Omega} (\mathcal{K}_{\mathcal{E}}(u)(x,y)u(x+y) \\&+ \alpha \mathcal{K}_{\mathcal{E}}(u)(x,y)\ln \mathcal{K}_{\mathcal{E}}(u)(x,y))dy,
    \end{split}
    \end{equation*}
    where 
        \[ \mathcal{K}_{\mathcal{E}}(u)(x,y) = \begin{cases}
        \frac{e^{-\frac{u(x+y)}{\alpha}}}{\int_\mathbb{B} e^{-\frac{u(x+z)}{\alpha}} dz}, \quad &y \in \mathbb{B}, \\
        0, &y \in \Omega \backslash \mathbb{B},
        \end{cases}
        \]
    is smooth erosion kernel.

    Through duality, we unify the form of smooth morphological operators. Please note that the above $\mathcal{K}_\mathcal{D}, \mathcal{K}_\mathcal{E}$ are classic softmax operator and softmin operator respectively, we can replace them with learnable kernel functions to implement more complex operations.
        
	Based on the above derivation, we can use smooth operators to approximate the morphological skeleton \eqref{skel}, and define smooth skeleton by
	\begin{equation*}
        \begin{split}
        \mathcal{S}^{\alpha}(u):=& \mathrm{Proj}_{[0,1]}\left(\sum\limits_{j=0}^J \mathcal{S}_j^{\alpha}(u)\right) \\
            =&ReLU\left(1-ReLU\left(1-\sum\limits_{j=0}^J \mathcal{S}_j^{\alpha}(u)\right)\right),
        \end{split}
	\end{equation*}
    \[\begin{split}
         \mathcal{S}_j^{\alpha}(u)=&\underbrace{\mathcal{E}^{\alpha}\circ\mathcal{E}^{\alpha}\circ\cdots\circ\mathcal{E}^{\alpha}}_j(u)- \mathcal{D}^{\alpha}(\mathcal{E}^{\alpha}(\underbrace{\mathcal{E}^{\alpha}\circ\cdots\circ\mathcal{E}^{\alpha}}_j(u)) \\=&\underbrace{\mathcal{E}^{\alpha}\circ\mathcal{E}^{\alpha}\circ\cdots\circ\mathcal{E}^{\alpha}}_j(u) - \mathcal{D}^{\alpha}\circ \underbrace{\mathcal{E}^{\alpha}\circ\cdots\circ\mathcal{E}^{\alpha}}_{j+1}(u).
    \end{split}\]
    The above projection operator $\mathrm{Proj}$  is to avoid numerical overflows after smoothing. Moreover, we have 
    \begin{ownprop}
        \label{th:var_s}
The variational of $\mathcal{S}^{\alpha}(u)$ with respect to $u$ is 
\begin{equation*}
\begin{array}{rl}
\frac{\delta \mathcal{S}^{\alpha}}{\delta u}
=&\mathcal{H}\left(1-ReLU\left(1-\sum\limits_{j=0}^J \mathcal{S}_j^{\alpha}(u)\right)\right)\\
&
\mathcal{H}\left(1-\sum\limits_{j=0}^J \mathcal{S}_j^{\alpha}(u)\right)
\displaystyle\biggl.\sum_{j=0}^J \frac{\delta \mathcal{S}_j^{\alpha} }{\delta u },
\end{array}
\end{equation*}
where $\mathcal{H}$ is the Heaviside step
function defined in \eqref{eq:Heaviside}, and
\[ \frac{\delta \mathcal{S}_j^{\alpha} }{\delta u } = \begin{cases}
    (1-\mathcal{K}_{\mathcal{D}}(e^{j+1})\mathcal{K}_{\mathcal{E}}(e^j))\prod\limits_{i=0}^{j-1}\mathcal{K}_{\mathcal{E}}(e^i), & j\ge1, \\
    1-\mathcal{K}_{\mathcal{D}}(e^{j+1})\mathcal{K}_{\mathcal{E}}(e^j),& j=0,
\end{cases}\]
\[e^i = \underbrace{\mathcal{E}^{\alpha}\circ\cdots\circ\mathcal{E}^{\alpha}}_{i}(u) ,\ i=1,2,\dots,\ \mathcal{K}_{\mathcal{E}}(e^0)=\mathcal{K}_{\mathcal{E}}(u).\]
\end{ownprop}
    
	Due to the smoothness of $\mathcal{E}^\alpha$ and $\mathcal{D}^\alpha$, it is easy to find that $\mathcal{S}^{\alpha}(u)$ is a smoothness of $\mathcal{S}(u)$. Figure \ref{fig:skelex} compares the proposed smooth skeleton with the classic morphological skeleton. Visually, the proposed skeleton is smoother, maintains better connectivity, and closely approximates the classical skeleton when $\alpha$ is sufficiently small.

    This key formula will be used in the model we establish below.
	
	\subsection{Segmentation Model with Morphological Skeleton Prior}
	Unlike methods that rely solely on loss functions such as \cite{shit2021cldice}, we aim to construct a variational segmentation model that enforces the segmentation results to preserve skeleton information throughout both training and prediction process. Firstly, assume $g$ is the ground truth segmentation of the sample $I$, let us define cost functional:
	\begin{equation}
		\label{skele}
		\mathcal{C}(u)=\frac{1}{2}\int_{\Omega}|\mathcal{S}^{\alpha}(u)(x)-\mathcal{S}^{\alpha}(g)(x)|^2dx ,
	\end{equation}
	where $\mathcal{S}^{\alpha}(u), \mathcal{S}^{\alpha}(g)$ are the skeletons of network output and ground truth $g$, respectively. This energy measures the difference between $\mathcal{S}^{\alpha}(u)$ and $\mathcal{S}^{\alpha}(g)$ and its variation with respect to $u$ can be written as
	\begin{equation*}
		\frac{\delta \mathcal{C}}{\delta u} = \frac{\delta\mathcal{C}}{\delta \mathcal{S}^{\alpha}}\cdot \frac{\delta \mathcal{S}^\alpha}{\delta u} 
			=(\mathcal{S}^{\alpha}(u)-\mathcal{S}^{\alpha}(g)) \cdot \frac{\delta \mathcal{S}^{\alpha}}{\delta u}.
	\end{equation*}
	
	Now, our segmentation model can be given as:
	\begin{equation}
		\label{MORSTD}
		\min\limits_{u\in[0,1]} \{ \langle-o,u\rangle+\gamma(\langle u,\ln u\rangle  +\langle 1-u, \ln(1-u) \rangle)+ \mathcal{R}(u)+\mathcal{C}(u) \},
	\end{equation}
	where $\mathcal{R}(u)=\lambda\langle u,f\ast(1-u)\rangle$ denotes the TD regularization term mentioned earlier in \eqref{TDterm}. By integrating the cost function \eqref{skele} directly into the variational model, we ensure that essential structural details are preserved throughout the optimization process of \eqref{MORSTD}.
	
	Since the fidelity term, regularization term and skeleton constraint term of optimization problem \eqref{MORSTD} all contain $u$, it is difficult to solve it directly using gradient methods. \textcolor{black}{The operator splitting method \cite{douglas1956numerical, ALM, Glowinski2016} is suitable for such objective functions. Combined with augmented lagrangian method \cite{glowinski1989augmented}, it decomposes a complex problem into two or more simple sub-problems by introducing new operators.} Here we introduce a new variable $w$ to transform the original problem into:
	\begin{equation*}
		\min\limits_{u\in[0,1]}  \mathcal{G}(u)+\mathcal{C}(w) \quad \mathrm{s.t.} \quad w-u = 0,
	\end{equation*}
	where $ \mathcal{G}(u)=\langle-o,u\rangle+\gamma(\langle u,\ln u\rangle  +\langle 1-u, \ln(1-u) \rangle) + \mathcal{R}(u)$. According to the penalty function method, we can get the object function:
	\begin{equation*}
		\min\limits_{u\in[0,1],w} \mathcal{G}(u)+\mathcal{C}(w) + \eta \|w-u\|_1,
	\end{equation*}
	Here, the main reason we choose the $L_1$ penalty term instead of $L_2$ is that the dual problem of $L_1$ is linear with respect to $u$, which allows the $u$ sub-problem to have a closed-form solution.
	
	\begin{ownprop}[Duality of $L_1$]
	    \label{th: dualL1}
		\[ \|y\|_1 = \|y\|_1^{**} = \max_{\|q\|_{\infty}\le 1}\langle q, y\rangle. \]
	\end{ownprop}
	\begin{IEEEproof}
        The detailed proof is shown in Appendix C.
    \end{IEEEproof}
    
	Now, we obtain an equivalent optimization problem:
	\begin{equation*}
		 \min\limits_{u\in[0,1],w}\max\limits_{\|q\|_{\infty}\le 1} \mathcal{G}(u)+\mathcal{C}(w) + \eta \langle q, w-u \rangle.
	\end{equation*}
	Then we can solve this problem by alternating optimization method to split it into the following three subproblems:
	\begin{equation*}
		\begin{cases}
			q^{t+1} = \arg\max\limits_{\|q\|_{\infty}\le 1}\{\langle q, w^t-u^t\rangle\},\\
			w^{t+1} = \arg\min\limits_{w}\{ \mathcal{C}(w)+\eta\langle q^{t+1}, w\rangle
			 \},  \\
			u^{t+1} = \arg\min\limits_{u\in[0,1]}\{ \mathcal{G}(u)-\eta\langle q^{t+1}, u\rangle \}.
		\end{cases}
	\end{equation*}
	We can approximately solve the first and second subproblems by projection gradient ascent and descent with one iteration and get:
	\begin{align*}
		q^{t+1} &\approx \mathrm{Proj}_{\|\cdot\|_{\infty}\le1}\{q^t+(w^t-u^t)\}, \\
		w^{t+1} &\approx w^t -\iota\left(\frac{\delta\mathcal{C}(w)}{\delta w}|_{w^t}+\eta q^{t+1}\right).
	\end{align*}
	For the third subproblem related to $u$, difference of convex functions (DC) algorithm can be employed to it by replacing concave $\mathcal{R}$ with its supporting hyperplane\cite{liu2022deep}, and we get a strictly convex optimization problem:
	\begin{equation*}
		u^{t+1} = \arg\min\limits_{u\in[0,1]}\begin{Bmatrix}
		\langle-o+p^t,u\rangle+\gamma\langle u,\ln u\rangle\\ + \gamma\langle 1-u, \ln(1-u) \rangle -\eta\langle q^{t+1}, u\rangle
		\end{Bmatrix},
	\end{equation*}
	Moreover, it has a closed-form solution of sigmoid function:
	\begin{equation*}
		\begin{split}
            u^{t+1} &= sigmoid\left(\frac{o-p^t+\eta q^{t+1}}{\gamma}\right) \\
            &=\left(1+exp(-\frac{o-p^t+\eta q^{t+1}}{\gamma})\right)^{-1},
        \end{split}
	\end{equation*}
	where $p^t = \lambda f * (1-2u^t) \in \delta \mathcal{R}(u^t)$ is subgradient of $\mathcal{R}(u)$.
	
	In summary, algorithm \ref{alg:STD} outlines the steps for solving the proposed model.
	
	\begin{algorithm}[H]
		\caption{Image segmentation with MorSP}\label{alg:STD}
		\begin{algorithmic}
			\STATE \textbf{Input:} segmentation feature $o$, skeleton label $ \mathcal{S}^{\alpha}(g) $
			\STATE \textbf{Output:} soft segmentation function $u$
			\STATE \textbf{Initialize:}
			\STATE \hspace{0.25cm} $u^0\leftarrow sigmoid(o)$,
			\STATE \hspace{0.25cm} $w^0 \leftarrow \frac{u^0+ \mathcal{S}^{\alpha}(g)}{2}$,
			\STATE \hspace{0.25cm} $q^0 \leftarrow \min(\max(w^0-u^0, -1), 1)$.
			\STATE \hspace{0.25cm} \textbf{for} $t=0,1,2,\dots$
			\STATE \hspace{0.5cm} \textbf{1.} update dual variable $q$:
			\[ \hat{q}^{t+1} \leftarrow q^t+(w^t-u^t), \]
			\[q^{t+1}\leftarrow \min(\max(\hat{q}^{t+1}, -1), 1).\]
			\STATE \hspace{0.5cm} \textbf{2.}  update variable $w$: 
			\[w^{t+1}\leftarrow w^t -\iota\left(\frac{\delta\mathcal{C}(w)}{\delta w}|_{w^t}+\eta q^{t+1}\right).\]
			\STATE \hspace{0.5cm} \textbf{3.} compute the solution by DC algorithm:
			\[ p^t \leftarrow \lambda f * (1-2u^t), \]
			\[ u^{t+1} \leftarrow sigmoid\left(\frac{o-p^t+\eta q^{t+1}}{\gamma}\right). \]
			\STATE \hspace{0.5cm} \textbf{4.} Convergence check.
			\STATE \hspace{0.25cm} \textbf{end}
		\end{algorithmic}
		\label{alg1}
	\end{algorithm}
 	
	\begin{figure}[!t]
		\centering
		\includegraphics[width=3.5in]{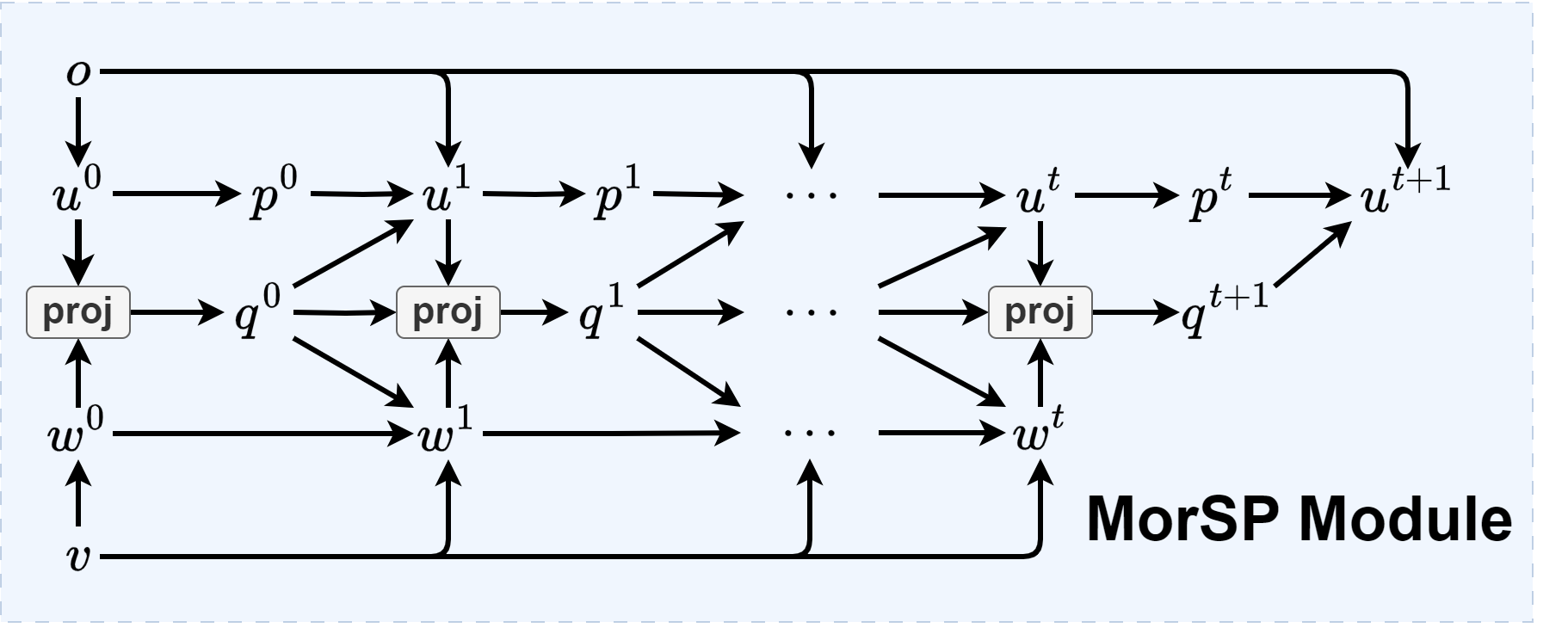}
		\caption{The network architecture of the proposed Morphological Skeleton Prior (MorSP) module.}
		\label{fig:MorSP}
	\end{figure}
 
	\subsection{SAM with Morphological Skeleton Prior (MorSP) Module}
	Next we will unroll the above algorithm \ref{alg:STD} into a network architecture called Morphological Skeleton Prior (MorSP) module that can be integrated into any segmentation neural network. Please note that in the variational model, the skeleton $\mathcal{S}^{\alpha}(g)$ is assumed to be known. In reality, during network prediction, such a strong prior cannot be obtained directly. Therefore, we introduce a token to learn soft skeleton label $v$. The architecture of MorSP module is illustrated in Fig.\ref{fig:MorSP}. Specifically, this module accepts two inputs: segmentation feature $o$ and soft skeleton label $v$ to compute the segmentation result $u$, and we just take $T=20$ layers to save computational sources.
	
    In order to integrate MorSP module into SAM, we add a soft skeleton token to mask decoder and replace the final decoding operator $\mathcal{H}$ in SAM \eqref{eq:sam}. So the proposed SAM-MorSP can be written as 
	\begin{equation}
		\label{SAM-MorSP}
		\begin{cases}
			F_I = \Phi_{\theta_1}(I),\ T_p, T_m = \Phi_{\theta_2}(p, m), \\
			o, v = \Phi_{\theta_3}(F_I + T_m, T_p),\\
			u^{\ast} = \arg\min\limits_{\substack{u\in[0,1], w}}\max\limits_{\|q\|_{\infty}\le 1}\mathcal{G}(u)+\mathcal{C}(w) + \eta \langle q, w-u \rangle.
		\end{cases}
	\end{equation}
	The third sub-equation in \eqref{SAM-MorSP} represents MorSP module, which endows SAM the property of preserving skeleton prior konwledge during prediction.
	
	\begin{figure}[!t]
		\centering
		\includegraphics[width=3.5in]{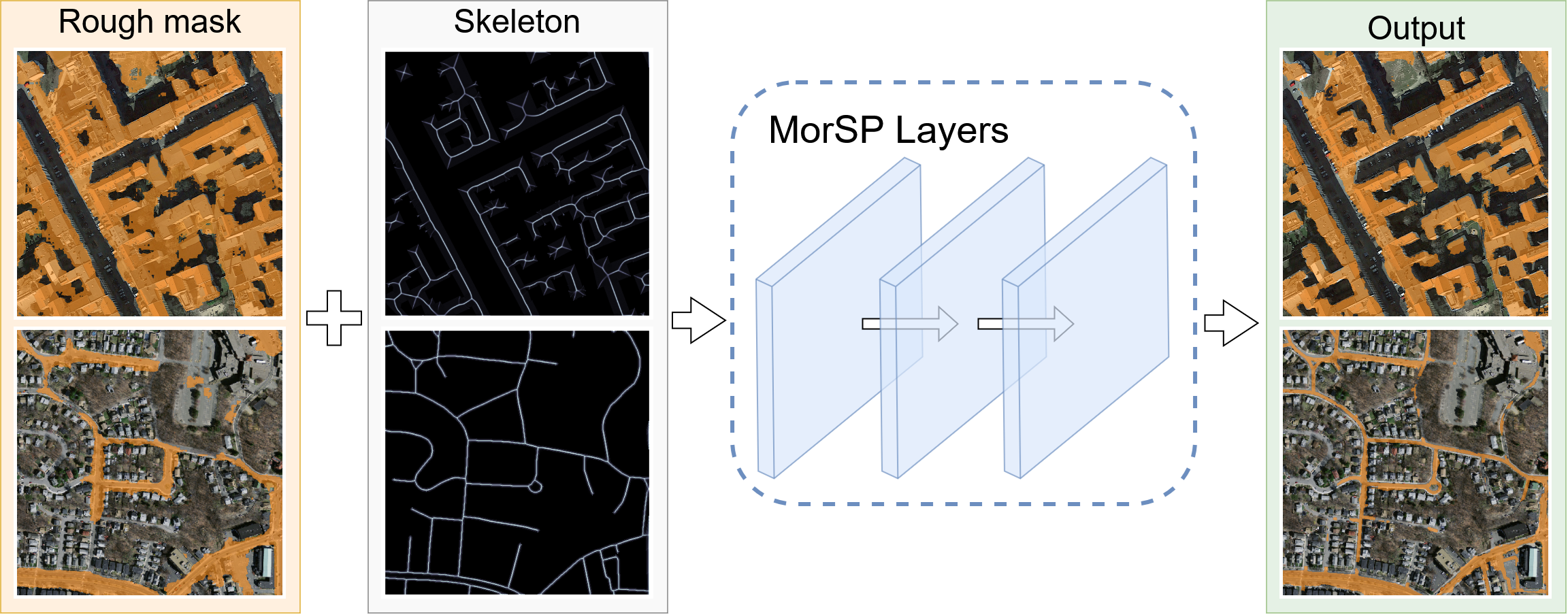}
		\caption{The performance of proposed MorSP module: the first column is the rough masks generated by networks, the second column is the input skeleton, and the last column is the output mask.}
		\label{fig:result1}
	\end{figure}
	
	\section{The Performance of MorSP Module}
	In this section, we design an experiment to evaluate the performance of the proposed Algorithm \ref{alg:STD} on some remote sensing images.
	 
    We select two images of buildings and roads, and use a segmentation network to generate low-quality, rough masks for these images. We searve these rough masks as the segmentation feature $o$ and the skeleton of ground truth as the soft skeleton $\mathcal{S}^{\alpha}(g)$ to post-process the rough masks. The hyper-parameters of MorSP module are all learnable and their initial values are shown in Table \ref{tab:MorSP}.
	
	The results are shown in Figure \ref{fig:result1}. Compared to original rough masks, the post-processed outputs from the MorSP module more effectively preserve structural features, such as connectivity and holes. This confirms that our method offers significant advantages in the segmentation of remote sensing images.
	
	\begin{table}[H]
		\caption{Initial hyper-parameters setting of MorSP module\label{tab:MorSP}.}
		\centering
		\begin{tabular}{ccccccc}
			\toprule[0.1em]
			size of $k$ & iterations & $\gamma$ & $\lambda$ &$\alpha$ & $\eta$ &  $\iota$  \\
			\midrule
			5 & 20 & 1 &1 & 0.05 & 1 & 1$\times 10^{-2}$  \\
			\bottomrule[0.1em]    
		\end{tabular}
	\end{table}
	
	\begin{figure*}[!t]
		\centering
		\includegraphics[width=0.9\textwidth]{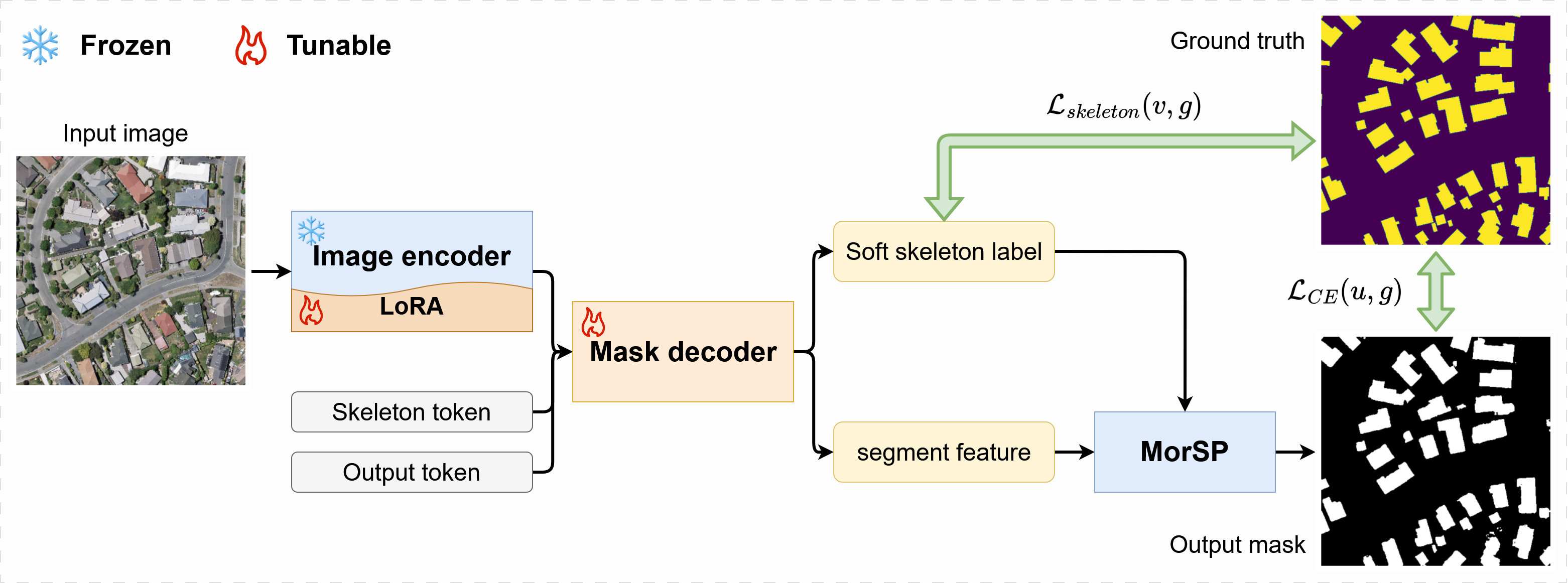}
		\caption{The network architecture of SAM-MorSP.}
		\label{fig:SAMMorSP}
	\end{figure*}
	
	\section{Numerical Experiment I: Results of SAM-MorSP}
	In this section, we will train SAM-MorSP on remote sensing datasets that include buildings, roads and water, and conduct ablation experiments to compare its performance with SAM trained using two other methods. The first method involves only fine-tuning the pretrained SAM model, and the second method employs fine-tuning with a modified skeleton loss function. The experimental results demonstrate that SAM-MorSP is superior in maintaining structural details and improving segmentation accuracy, especially in preserving the geometric integrity of remote sensing images.
	
	\subsection{Training strategy and experimental setup}
	SAM provides three scales of image encoders: ViT-B, ViT-L, and ViT-H, with parameter sizes ranging from 91M to 636M. Notably, ViT-B achieves over 90\% of the performance of ViT-H \cite{kirillov2023segment}. Thus, we select ViT-B as our encoder backbone to reduce computational demands.
	
	To retain SAM's original generalization ability while efficiently fine-tuning it for remote sensing datasets, we freeze all image encoder's parameters and introduce LoRA layers for fine-tuning it. Then we replace the decoder operator with MorSP module as the final layers, so the decoder's parameters are all tunable .The architecture of the entire network is illustrated in Figure \ref{fig:SAMMorSP}.
	
	The LoRA layers are applied to the Query (Q) and Value (V) components of the transformer blocks in image encoder, with each layer consisting of two matrices of rank 4. The MorSP module's hyperparameters are all learnable and their initial values are shown in Table \ref{tab:MorSP}. As a result, SAM-MorSP contains around 4M trainable parameters (3.9M for mask decoder, 147K for LoRA). 
 
    The loss functions employed during training are binary cross-entropy loss 
	\begin{equation*}
		\mathcal{L}_{CE}(u, g) = - \langle g, \log(u) \rangle - \langle (1-g), \log(1-u) \rangle,
	\end{equation*}
	and cl-dice loss\cite{shit2021cldice}
	\begin{equation*}
		\mathcal{L}_{skeleton}(v, g) = 1- 2\times \frac{T_{prec}(v,g)\times T_{sens}(v,g)}{T_{prec}(v,g)+ T_{sens}(v,g)},
	\end{equation*}
	where
	\[ T_{prec} = \frac{\langle g, \mathcal{S}(v)\rangle}{\|\mathcal{S}(v)\|_1 }, \quad T_{sens} = \frac{\langle \mathcal{S}(g), v \rangle }{\|\mathcal{S}(g)\|_1 }. \]
	We train the model with the weighted loss function:
	\begin{equation*}
	    \mathcal{L}(u, v, g) = (1-\lambda_l) \mathcal{L}_{CE}(u,g) + \lambda_l \mathcal{L}_{skeleton}(v,g).
	\end{equation*}
	where $\lambda_l\in(0,1)$. We utilize this loss function to fine-tune SAM with $\lambda_l$ set to 0.1 during training. For comparison, we fine-tune SAM using two different methods. The first method, referred to as SAM-fine, retains the original decoder operator $\mathcal{H}$ and employs only cross-entropy loss. The second method maintains the same network  as the first but utilizes a different loss function:
	\begin{equation*}
		\mathcal{L}(u, g) = (1-\lambda_l) \mathcal{L}_{CE}(u,g) + \lambda_l \mathcal{L}_{skeleton}(u,g),
	\end{equation*}
	and the model is denoted as SAM-cl. We train these models on an 32GB NVIDIA Tesla V100 GPU by AdamW optimizer using the same learning strategy. Besides, the batch size is set to 1 and the initial learning rate is set to $2\times10^{-4}$.

    
    \begin{figure*}[ht]
		\centering
		\includegraphics[width=0.8\textwidth]{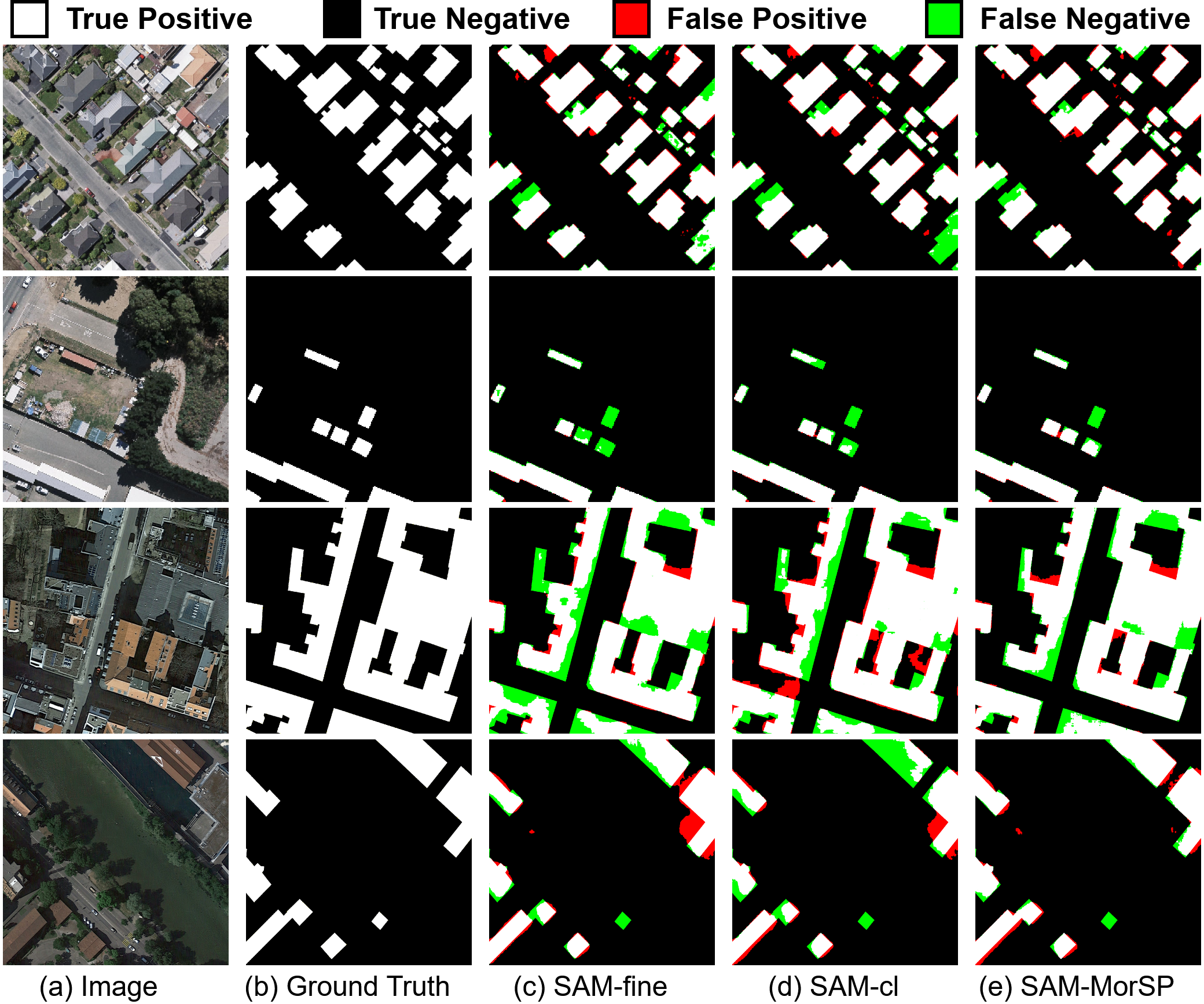}
		\caption{Visual result of SAM-fine, SAM-cl, SAM-MorSP on WHU and LAIS building test sets.}
		\label{fig:building}
	\end{figure*}
    
	\subsection{Datasets}
	The proposed model is fine-tuned and evaluated on remote sensing datasets consisting of three types of objects: buildings, roads and water, all of which contain thousands of aerial or satellite images. We apply simple preprocessing to the datasets, which is shown in Table \ref{tab:dataset}.
	
	\begin{table}[H]
		\caption{Data preprocessing of the datasets used \label{tab:dataset}.}
		\centering
            \resizebox{\linewidth}{!}{
		\begin{tabular}{cccc}
			\toprule[0.1em]
			Dataset & Type &Image size & Train-Validate-Test \\
			\midrule[0.05em]
			WHU & building &512$\times$512 &  4736-1036-2416 \\
			LAIS & building \& road &3000+- $\rightarrow$ 1024&  2246-562-1240 \\
			Massachusetts & road &1500 $\rightarrow$ 500 & 2000-126-441  \\
			\textcolor{black}{WHU-OPT-SAR} & \textcolor{black}{water} & \textcolor{black}{5556$\times$3704 $\rightarrow$ 926} & \textcolor{black}{1800-240-360} \\
			\bottomrule[0.1em]    
		\end{tabular}}
	\end{table}

	\subsubsection{WHU Building Dataset \cite{ji2018fully}}
	The dataset was created by Wuhan University and contains manually extracted samples of buildings from aerial and satellite images. For our experiments, we selected the aerial image subset, which consists of 8,189 images, each with a resolution of $512 \times 512$ pixels. These images were captured from Christchurch, New Zealand, and include over 220,000 individual buildings. It is widely used to train building extraction models due to its high-quality labels.
	
	\subsubsection{Learning Aerial Image Segmentation from Online Maps (LAIS) Dataset \cite{kaiser2017learning}} The LAIS dataset contains high-resolution aerial images of five cities from Google Maps, combined with pixel-level labels for buildings, roads and background from OpenStreetMap (OSM). Since OSM data is generated by volunteers, these labels inevitably contain some obvious errors with a offset as large as 10 pixels. Therefore, it can be used for for evaluating the performance of deep learning models in the presence of noisy data. We choose images from Berlin, Chicago, Zurich and Potsdam and randomly divided 30\% of them into the test set for our experiment. Due to the large image size, each original images is cropped to four images resized to 1024$\times$1024 for training.
	
	\subsubsection{Massachusetts Road  Dataset\cite{MnihThesis}}
	The Massachusetts Roads dataset consists of 1,171  aerial color images with 1500 $\times$ 1500 pixels, covering urban, rural, and mountainous areas of Massachusetts, USA. The images have a relatively low resolution of 1 m and there are some visible annotation errors in labels such as incomplete road markings. We cut each image into 500$\times$500 patches and selected 2000 of them with more roads for the training set, and adopted the validate set and test set officially divided.

    \textcolor{black}{
        \subsubsection{WHU-OPT-SAR Dataset\cite{LI2022102638}} This dataset contains 100 RGB, Near Infrared (NIR) optical images  and Synthetic Aperture Radar (SAR) images with a resolution of 5556$\times$3704 pixels in Hubei Province, China. In this experiment, we selected the optical images to segment the water. Each image is cropped into 24 patches with 926$\times$926 pixels, and the RGB channel is input into the encoder, and the NIR channel is used as mask prompt $m$ to input into the decoder.
    }
    
    \begin{figure*}[!ht]
		\centering
		\includegraphics[width=0.8\textwidth]{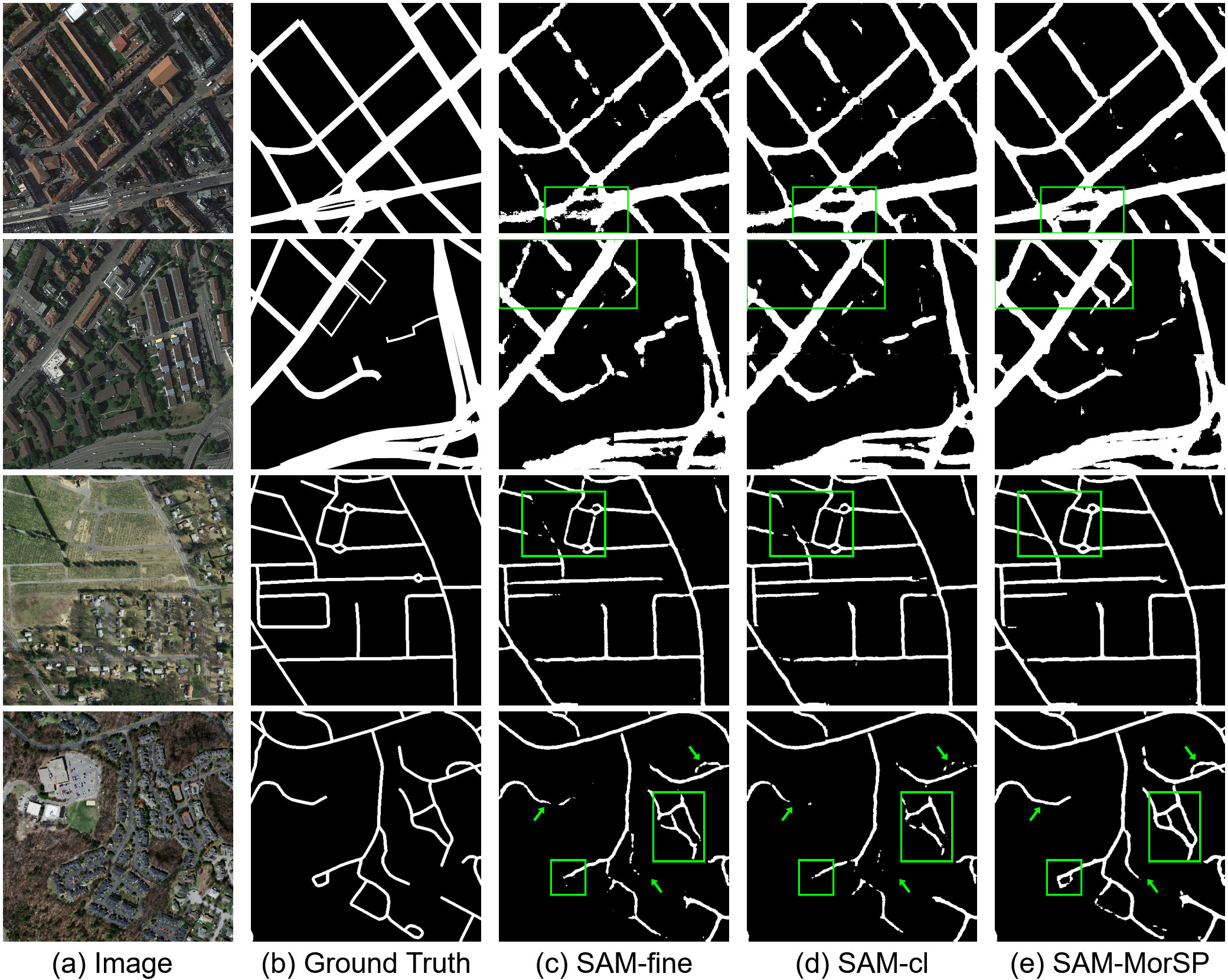}
		\caption{Visual result of SAM-fine, SAM-cl, SAM-MorSP on LAIS and Massachusetts roads test sets.}
		\label{fig:road}
	\end{figure*}
	
    \begin{table*}[!ht]
		\caption{Comparison of Segmentation Metrics on Building and Road datasets\label{tab:result}.}
		\centering
		\begin{threeparttable}
			\begin{tabular}{c|cc|cccc|cc|cccc}
				\toprule[0.1em]
				Type &Dataset & Model & $F_1$ $\uparrow$ & IOU $\uparrow$& Pre $\uparrow$& Rec $\uparrow$& Dataset &Model & $F_1$ $\uparrow$ & IOU $\uparrow$ & Pre$\uparrow$ & Rec $\uparrow$  \\
				\midrule[0.05em]
				\multirow{3}{*}{Building}&\multirow{3}{*}{WHU} & SAM-fine& 92.32 &87.85	&93.30& 93.63 & \multirow{3}{*}{LAIS} & SAM-fine&80.60&69.50&\textbf{83.39}&80.14 \\
				& &  SAM-cl & 92.96&	88.54&	93.84&93.79 & & SAM-cl &80.54 &69.16 &82.09 &81.23 \\
				& &  SAM-MorSP & \textbf{93.77}&\textbf{89.34} &	\textbf{94.46}&\textbf{94.01} & &SAM-MorSP & \textbf{81.39} & \textbf{70.12} &80.10 & \textbf{84.70} \\
				\midrule[0.05em]
				\multirow{3}{*}{Road} & \multirow{3}{*}{Massachusetts} & SAM-fine &74.14 & 60.73 & \textbf{79.27} & 71.53 & \multirow{3}{*}{LAIS} &SAM-fine & 75.33 & 63.35 & 77.09 & 76.62 \\
				& & SAM-cl & 74.67 & \textbf{61.38} & 79.15 & 72.67 & & SAM-cl & 74.68 & 62.68 & 76.62 & 76.04 \\
				& & SAM-MorSP & \textbf{74.98} & 61.24 & 77.60 & \textbf{74.33} & & SAM-MorSP & \textbf{76.24} & \textbf{63.57} & \textbf{78.59} & \textbf{76.79} \\
				\bottomrule[0.1em]    
			\end{tabular}
			\begin{tablenotes}
				\item[*] SAM-fine refers to the fine-tuned Segment Anything Model, and SAM-cl refers to Segment Anything Model trained with the cl-dice loss, and SAM-MorSP refers to proposed model (See Figure\ref{fig:SAMMorSP}). Three models have the same number of learnable parameters.
			\end{tablenotes}
		\end{threeparttable}
	\end{table*}
	

    \begin{figure*}[!ht]
		\centering
		\includegraphics[width=0.8\textwidth]{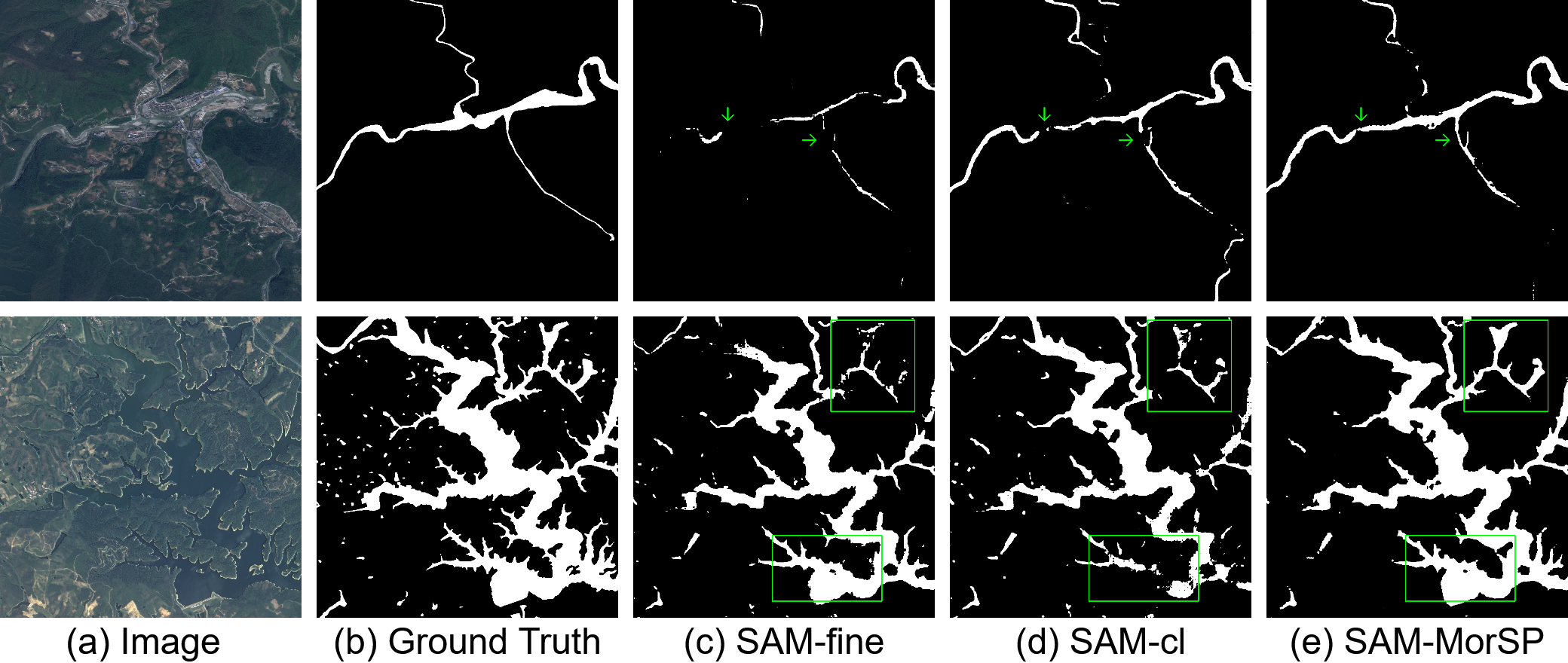}
		\textcolor{black}{\caption{Visual result of SAM-fine, SAM-cl, SAM-MorSP on LAIS and WHU-OPT-SAR test set.}\label{fig:water}}
		
	\end{figure*}

        \subsection{Performances and Analyses}
	The performance of above models are evaluated by four metrics including $F_1$ score, intersection over union (IOU), precision (Pre), recall (Rec). We use $\Omega_u, \Omega_g \subset \Omega$ to denote the segmentation area of the network prediction and the ground truth respectively, and the equations of evaluation metrics are as follows:
	\begin{equation*}
        F_1 = 2\times \frac{|\Omega_u\cap\Omega_g|}{|\Omega_u|+|\Omega_g|}, \quad
		IOU = \frac{|\Omega_u\cap\Omega_g|}{|\Omega_u\cup\Omega_g|},
	\end{equation*}
	\begin{equation*}
		Pre =  \frac{|\Omega_u\cap\Omega_g|}{|\Omega_u|}, \qquad	Rec =  \frac{|\Omega_u\cap\Omega_g|}{|\Omega_g|},
	\end{equation*}
	where $|\Omega_u|$ denotes the number of pixels contained in $\Omega_u$. Obviously, all these metrics range from 0 to 1, with higher values indicating better segmentation performance.

    \begin{table}[htb]
        \centering
        \textcolor{black}{
            \caption{Comparison of Segment metrics on WHU-OPT-SAR dataset}\label{tab:water}
                \begin{tabular}[c]{c|cccc}
                    \toprule[0.1em]
                    Model
                    & $F_1$ \textcolor{black}{$\uparrow$} & IOU \textcolor{black}{$\uparrow$} & Pre \textcolor{black}{$\uparrow$} & Rec \textcolor{black}{$\uparrow$} \\
                    \midrule[0.05em]
                    SAM-fine &  61.24 & 50.35 & \textbf{79.22} & 57.07 \\
                    SAM-cl & 59.54 & 47.77 & 65.71 & \textbf{64.60} \\
                    SAM-MorSP & \textbf{62.06} & \textbf{50.94} & 72.03 & 62.82 \\
                    \bottomrule[0.1em]
                \end{tabular}
        }
    \end{table}
	
	The segmentation performance of the three fine-tuned models on different datasets is summarized in Table \ref{tab:result}, \ref{tab:water}.  Notably, after fine-tuning only about 4\% of the parameters of the original SAM ViT-B, all three models demonstrate strong segmentation capabilities on remote sensing datasets.
	
	Incorporating skeleton constraint in loss function during training allows SAM-cl to achieve superior results on both the WHU and Massachusetts datasets, particularly in capturing fine targets and preserving structural details. This method effectively maintains intricate features and enhances the model's ability to delineate complex shapes with high precision. However, the cl-dice loss function faces challenges on the LAIS and WHU-OPT-SAR dataset, which contains noise and label offsets, highlighting the limited stability of such methods rely on loss function, especially when applied to datasets with lower-quality annotations.
	
	The proposed SAM-MorSP model demonstrates superior performance across all datasets, with an approximate 2\% improvement in recall compared to SAM-fine. This improvement stems from two key advantages. First, SAM-MorSP excels at detecting small buildings, which are often challenging to identify, especially in cluttered or complex environments  (See Figure \ref{fig:building}, where the white means true positive, black means true negative, red means false positive and green means false negative). Additionally, it performs better in segmenting objects with intricate shapes, ensuring accurate delineation of complex structures.  Second, SAM-MorSP significantly enhances road and water segmentation by better preserving road connectivity (see Figure \ref{fig:road}, \ref{fig:water}), leading to more precise extraction of complex road features like intersections, roundabouts and slender rivers. These strengths make SAM-MorSP particularly well-suited for remote sensing tasks.

    \textcolor{black}{
        \subsection{Optimal loss function weight}
         The weight $\lambda_l$ in the loss function is a critical hyper-parameter for model training, so we conduct additional experiments with two different $\lambda_l$ on the Massachusetts road dataset to select the optimal weight. The results are presented in Table \ref{tab:weight}, which suggest that the MorSP model consistently outperforms the baseline and as long as an appropriate weight $\lambda_l$ is selected and 0.1 is likely the optimal weight.
    }
    \begin{table}[htbp]
			\centering
                \textcolor{black}{
			    \caption{Results of different loss function weights.}
			    \label{tab:weight}
                \begin{tabular}[c]{c|cccc}
                    \toprule[0.1em]
                    Weight $\lambda_l$ 
                    & $F_1$ \textcolor{black}{$\uparrow$} & IOU \textcolor{black}{$\uparrow$} & Pre \textcolor{black}{$\uparrow$} & Rec \textcolor{black}{$\uparrow$}  \\
                    \midrule[0.05em]
                    0.05 &  74.1 & \textbf{61.45} & 79.87 & 72.56 \\
                    0.1 & \textbf{74.98} & 61.24 & 77.60 & \textbf{74.33} \\
                    0.2 & 74.13 & 61.44 & \textbf{80.21} & 72.60 \\
                    \bottomrule[0.1em]
                \end{tabular}}
	\end{table}
	
	\textcolor{black}{
    \subsection{Information of the Computational Demands}
    To finetune SAM, we incorporate two lightweight modules, LoRA\cite{hu2021lora} and MorSP. We evaluate the impact of these two modules on computational demands during segementing an image with $1024\times1024$ pixels. The results are presented in Table \ref{tab:comp}.
    }
    \begin{table}[htb]
        \centering
        \textcolor{black}{
            \caption{computational demands of different models}\label{tab:comp}
                \begin{tabular}{c|ccc}
                    \toprule[0.1em]
                    Model & Params \textcolor{black}{$\downarrow$} & Gflops \textcolor{black}{$\downarrow$} & inference time (ms) \textcolor{black}{$\downarrow$} \\
                    \midrule[0.05em]
                    SAM (vit-b) & 93.74M & 976.22 & 153.63 \\
                    + LoRA & 93.88M & 977.58 & 156.25 \\
                    + LoRA \& MorSP & 93.88M & 987.58 & 228.17 \\
                    \bottomrule[0.1em]
                \end{tabular}
        }
    \end{table}
    
    \begin{figure}[htbp]
		\centering
		\includegraphics[width=3.5in]{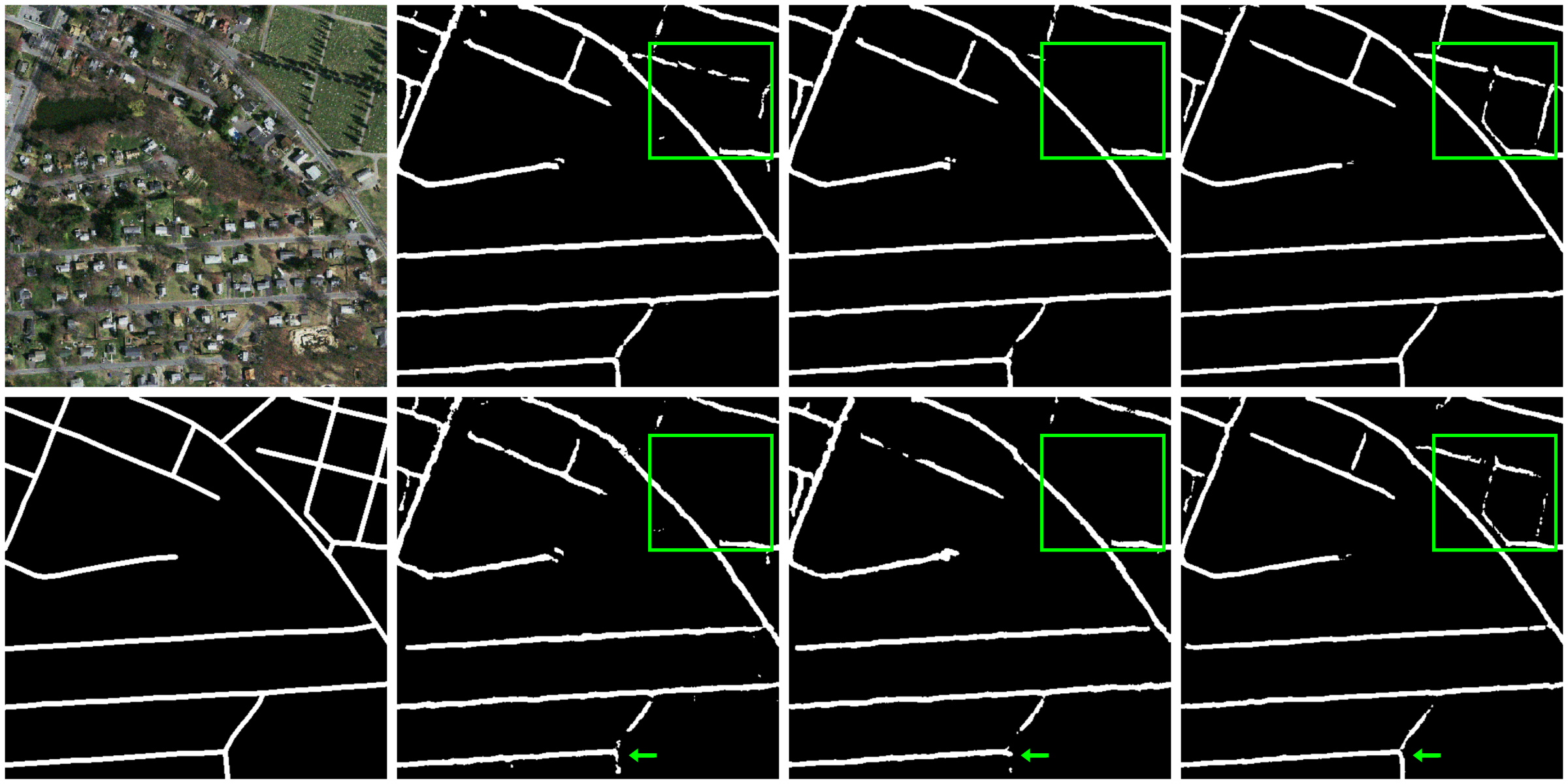}
		\textcolor{black}{\caption{Visual result of noise test. The first row (from left to right): noisy image, mask of SAM-fine, SAM-cl, SAM-MorSP on clean image. The second row: ground truth, mask of SAM-fine, SAM-cl, SAM-MorSP on noisy image.\label{fig:road_noisy}}}
	\end{figure}
	
	\subsection{Generalization Ability for Noise}
	To evaluate the robustness of the proposed model on noisy data, we added two levels of Gaussian noise to the test set images from the Massachusetts road dataset, with standard deviations of 0.05 and 0.1 (image gray scale range [0,1]). The results (See Table \ref{tab:noisy}) indicate that all three models experience performance degradation as noise levels increase, particularly under heavy noise. The added noise obscures road features, making it challenging for the network to accurately recognize portions of the road.
	
	While SAM-cl experiences a significant decline in performance, falling below the baseline SAM-fine on heavily noisy images, the proposed SAM-MorSP model consistently outperforms both. This is attributed to the TD regularization term and smooth morphological operators, which help to smooth the image by removing small noise while preserving major structural features. Figure \ref{fig:road_noisy}  illustrates the comparative performance of the three methods.
	
	\begin{table}[htb]
		\caption{Test results on noisy image on  Massachusetts road dataset\label{tab:noisy}.}
		\centering
        \resizebox{\linewidth}{!}{
		\begin{tabular}{cc|cccc}
			\toprule[0.1em]
			Standard deviation & Model & $F_1$ $\uparrow$ & IOU $\uparrow$& Pre $\uparrow$& Rec $\uparrow$ \\
			\midrule[0.05em]
			\multirow{3}{*}{0.05} & SAM-fine & 67.34 & 52.30 & 71.97 & 65.89 \\
			& SAM-cl & 67.46 & 52.63 & 69.45 & 68.67 \\
			& SAM-MorSP & \textbf{70.10} & \textbf{55.61} & \textbf{73.80} & \textbf{69.05} \\
			\midrule[0.05em]
			\multirow{3}{*}{0.1} & SAM-fine & 48.35 & 35.27 & 70.32 & 40.71 \\
			& SAM-cl &44.43 & 32.72 & 72.43 &37.46 \\
			& SAM-MorSP & \textbf{52.30} & \textbf{40.31} &\textbf{77.04} & \textbf{44.79} \\
			\bottomrule[0.1em]    
		\end{tabular}}
	\end{table}

\begin{figure}[htb]
			\centering
			\includegraphics[width=3.5in]{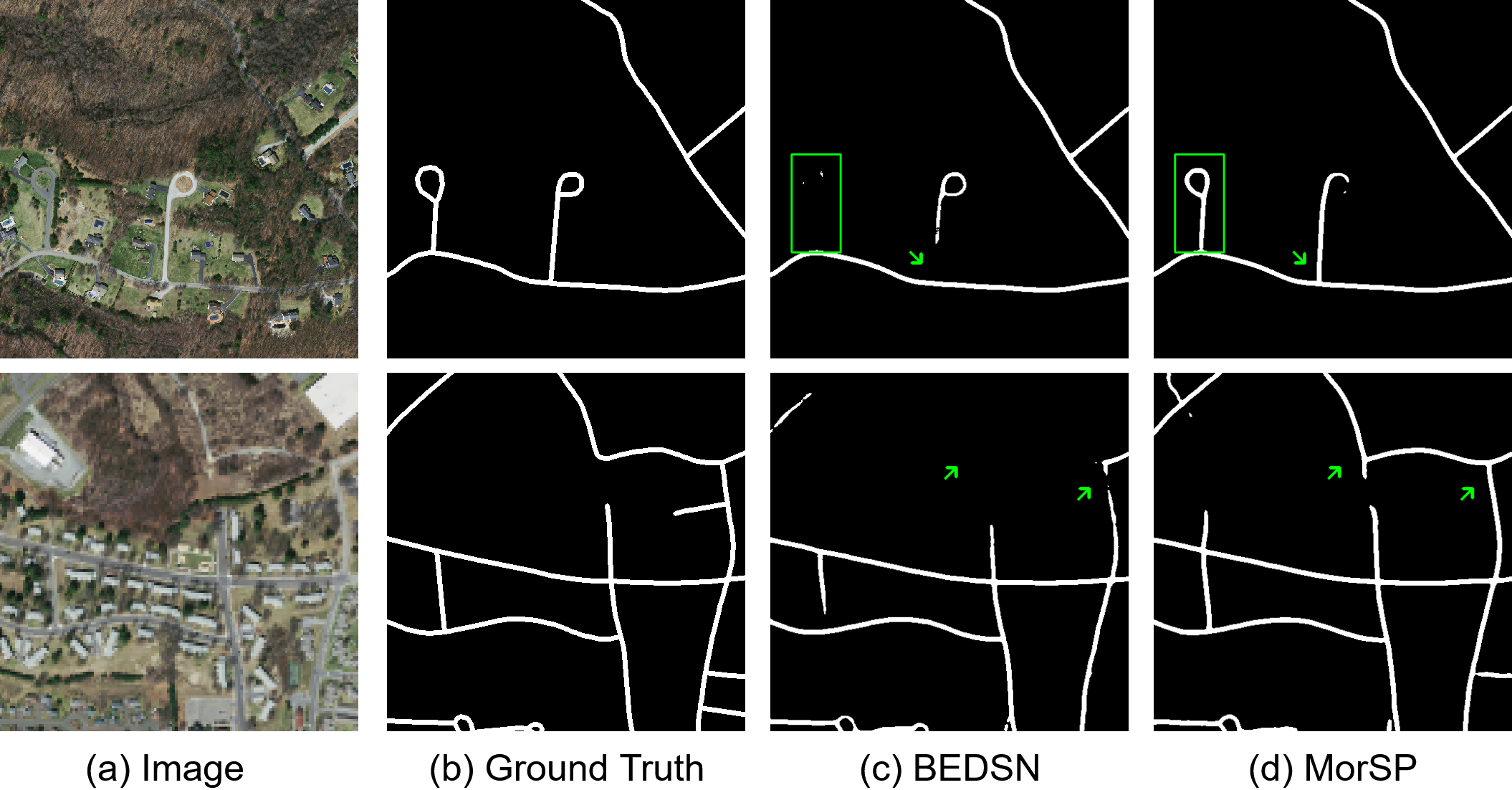}
			\textcolor{black}{\caption{Visual result of BEDSN on Massachusetts road dataset.}\label{fig:bedsn-road}}
		\end{figure}

    \textcolor{black}{
    \section{Numerical experiment II: MorSP on other backbone}
    The proposed MorSP module is based on variational model and can be integrated into any existing segmentation network. We select the Boundary-enhanced Dual-stream Network (BEDSN)\cite{li2024boundary} as the backbone with a convolutional layer (with 1600 parameters) to learn the skeleton prior $v$ and integrate the MorSP module at the end of the network. And the result of corresponding numerical experiments on WHU building and Massachusetts road datasets are shown in Table \ref{tab:BEDSN} and Figure \ref{fig:bedsn-road}.
    }
    
    \begin{table}[htb]
        \centering
        \textcolor{black}{\caption{Segmentation Metrics of numerical experiment on BEDSN}
        \label{tab:BEDSN}
        \begin{tabular}{cc|cccc}
            \toprule[0.1em]
            Dataset & Model & $F_1$ \textcolor{black}{$\uparrow$} & IOU \textcolor{black}{$\uparrow$} & Pre \textcolor{black}{$\uparrow$} & Rec \textcolor{black}{$\uparrow$}  \\
            \midrule[0.05em] 
            \multirow{2}{*}{WHU} & BEDSN & 91.16 & 87.12 & 93.41 & 92.81 \\
            & + MorSP & \textbf{92.42} & \textbf{89.00} & \textbf{94.83} & \textbf{93.58} \\
            \midrule
            \multirow{2}{*}{Massachusetts} & BEDSN & 79.50 & 59.93 & \textbf{89.19} & 75.73 \\
            & +MorSP & \textbf{81.20} & \textbf{61.84} &  86.50 & \textbf{80.21} \\
            \bottomrule[0.1em]
        \end{tabular}}
    \end{table}
    
	\section{Conclusion and Discussion}
    Objects in remote sensing images exhibit distinct geometric characteristics, featuring fine local structures and complex overall shapes. While Vision-Transformers (ViT) pretrained on large image datasets are powerful semantic feature extractors, they often struggle with small targets and intricate details. Recognizing that morphological skeletons can effectively describe the geometric structure and topological features of these objects, we propose a variational segmentation algorithm that integrates smooth morphological skeleton priors and unrolls it into the architecture of SAM. The novel fusion model is mathematically interpretable and constrains outputs to preserve the learned skeleton prior during both training and prediction.

    The proposed method only modifies the final activation layer of the network, demonstrating superior robustness against noisy data compared to existing methods that rely on modified loss functions. This adaptability facilitates easy integration with other data-driven models, enabling the segmentation of various complex structures.

    \section*{Appendix A \\ Proof of Proposition \ref{th:SM}} 
    We remark that $\mathbb{M}=\{x+y: y = \arg\max_{z\in\mathbb{B}}u(x+z)\} $ and take $\hat{x} \in \mathbb{M}$, then
    \[   
    \begin{split}
		\mathcal{D}^{\alpha}(u)(x) &= u(\hat{x}) + \alpha\ln(\int_{z\in \mathbb{B}}e^{\frac{u(x+z)-u(\hat{x})}{\alpha}}dz) \\
		&\le u(\hat{x}) + \alpha \ln (|\mathbb{B}|),
	\end{split}
    \]
     and we have $\varlimsup\limits_{\alpha\rightarrow 0^{+}}\mathcal{D}^{\alpha}(u)(x) \le u(\hat{x}). $
    Using the definition of continuity of $u$ at $\hat{x}$, one can prove that $\forall \epsilon>0,
     \varliminf\limits_{\alpha\rightarrow 0^{+}}\mathcal{D}^{\alpha}(u)(x) \ge u(\hat{x}) - \epsilon. $ From the arbitrariness of $\epsilon$, we get
     $\varliminf\limits_{\alpha\rightarrow 0^{+}}\mathcal{D}^{\alpha}(u)(x) \ge u(\hat{x})$. Therefore,
     $$\varlimsup\limits_{\alpha\rightarrow 0^{+}}\mathcal{D}^{\alpha}(u)(x)=\varliminf\limits_{\alpha\rightarrow 0^{+}}\mathcal{D}^{\alpha}(u)(x)=u(\hat{x}).$$
     It implies that
      $$\lim\limits_{\alpha\rightarrow0^{+}} \mathcal{D}^{\alpha}(u)(x) =u(\hat{x})=\max_{z\in \mathbb{B}} u(x+z).$$

    \section*{Appendix B \\ Proof of Proposition \ref{th:SD}}
    According to the definition of Fenchel-Legendre transformation, for a fixed point $x\in\Omega$,
    \[
    \begin{split}
    (\mathcal{D}^{\alpha})^ \ast (k)(x)  :=& \max\limits_{u}\{ \langle k, u \rangle_\mathbb{B} - \mathcal{D}^{\alpha}(u)(x) \} \\
     =& \begin{cases}
    \alpha\langle k, \ln k\rangle_\mathbb{B}, &k \in \mathbb{K}(x), \\
    +\infty, &else,
    \end{cases}
    \end{split}
    \]
    where $\mathbb{K}(x) = \{k:\Omega\rightarrow[0,1],\ \int_\mathbb{B} k(x+z)dz=1 \},\ \forall x\in\Omega$. Therefore, we have
    \[  
    \begin{split}
    (\mathcal{D}^{\alpha})^{\ast\ast}(u)(x) :&= \max\limits_{k} \{ \langle k, u \rangle_\mathbb{B} - (\mathcal{D}^{\alpha})^\ast k(x)  \} \\
    &= \max\limits_{k\in\mathbb{K}(x)} \{ \langle k, u \rangle_\mathbb{B} - \alpha\langle k, \ln k\rangle_\mathbb{B}  \},
    \end{split}
    \]
    and the above problem has a closed form solution:
    \[
    \hat{k}(y) = \frac{e^{\frac{u(x+y)}{\alpha}}}{\int_\mathbb{B} e^{\frac{u(x+z)}{\alpha}}dz}.
    \]
    
    \section*{Appendix C  \\ Proof of Proposition \ref{th: dualL1}}
    We remark that $h(y) = \|y\|_1$, and its French-Legendre transform is
    \[ \begin{split}
        h^{\ast}(q) &= \max\limits_{y} \{ \langle q, y \rangle - h(y) \},\\
        &= \begin{cases}
            +\infty, \quad & \|q\|_{\infty}>1, \\
            0, \quad & \|q\|_{\infty}\le 1.
        \end{cases}
    \end{split}  \]
    Then we have the dual norm of $L^1$:
    \[ \begin{split}
        h^{\ast\ast}(y) &= \max\limits_{q}\{\langle y,q \rangle - h^{\ast}(q) \} = \max\limits_{\|q\|_{\infty}\le 1} \langle y,q \rangle.
    \end{split}
    \]
	
	\bibliographystyle{IEEEtran}
	\bibliography{reference.bib}

    \begin{IEEEbiography}[{\includegraphics[width=1in, height=1.25in, clip, keepaspectratio]{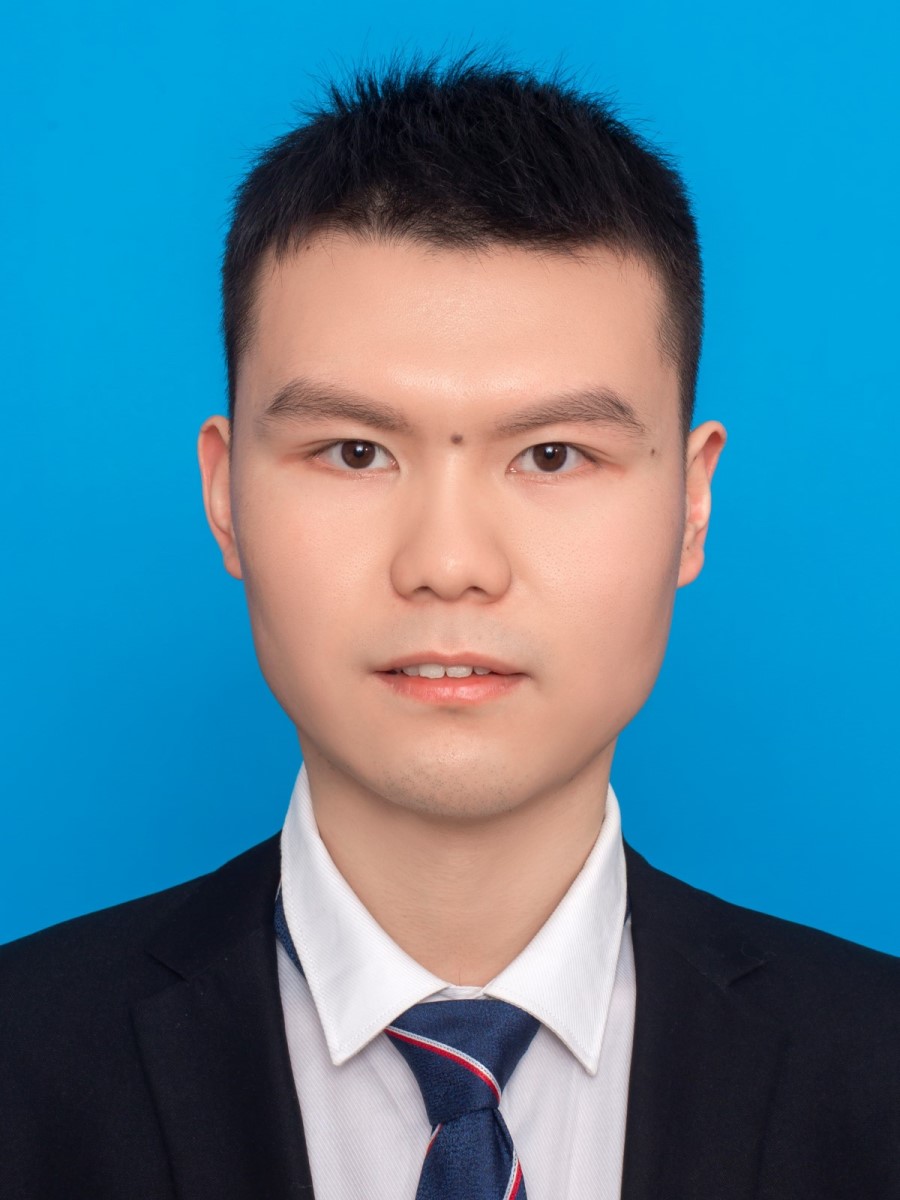}}]{Jun Xie}
        received the B.S. degree in mathematics from the School of Mathematical Sciences at Beijing Normal University in 2023. He is currently pursuing the M.S. degree at Beijing Normal University. His research interests include variational image process, deep learning and its applications in remote sensing image.
    \end{IEEEbiography}
    \begin{IEEEbiography}[{\includegraphics[width=1in, height=1.25in, clip, keepaspectratio]{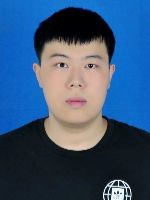}}]{Wenxiao Li}
        received the B.S. degree in Information and Computing Science from Hunan Normal University, Hunan, China, in 2020, and the M.S. degree in computational mathematics from Beijing Normal University, Beijing, China, in 2024, where he is currently pursuing the Ph.D. degree in computational mathematics. His main research interests include variational image processing, computer vision and deep learning.
    \end{IEEEbiography}
    \begin{IEEEbiography}[{\includegraphics[width=1in, height=1.25in, clip, keepaspectratio]{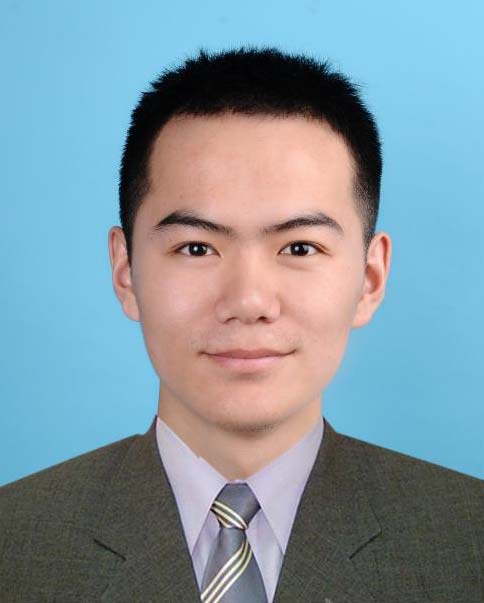}}]{Faqiang Wang}
        received the B.S. degree in mathematics and applied mathematics from Heilongjiang University, China in 2014. He received the M.S. degree in computational mathematics and the Ph.D. degree in applied mathematics from Beijing Normal University (BNU), China in 2017 and 2020 respectively. He is currently a lecturer at BNU. His main research interests include variational and PDE-based image processing, optimal transport and deep learning based methods.
    \end{IEEEbiography}
    \begin{IEEEbiography}[{\includegraphics[width=1in, height=1.25in, clip, keepaspectratio]{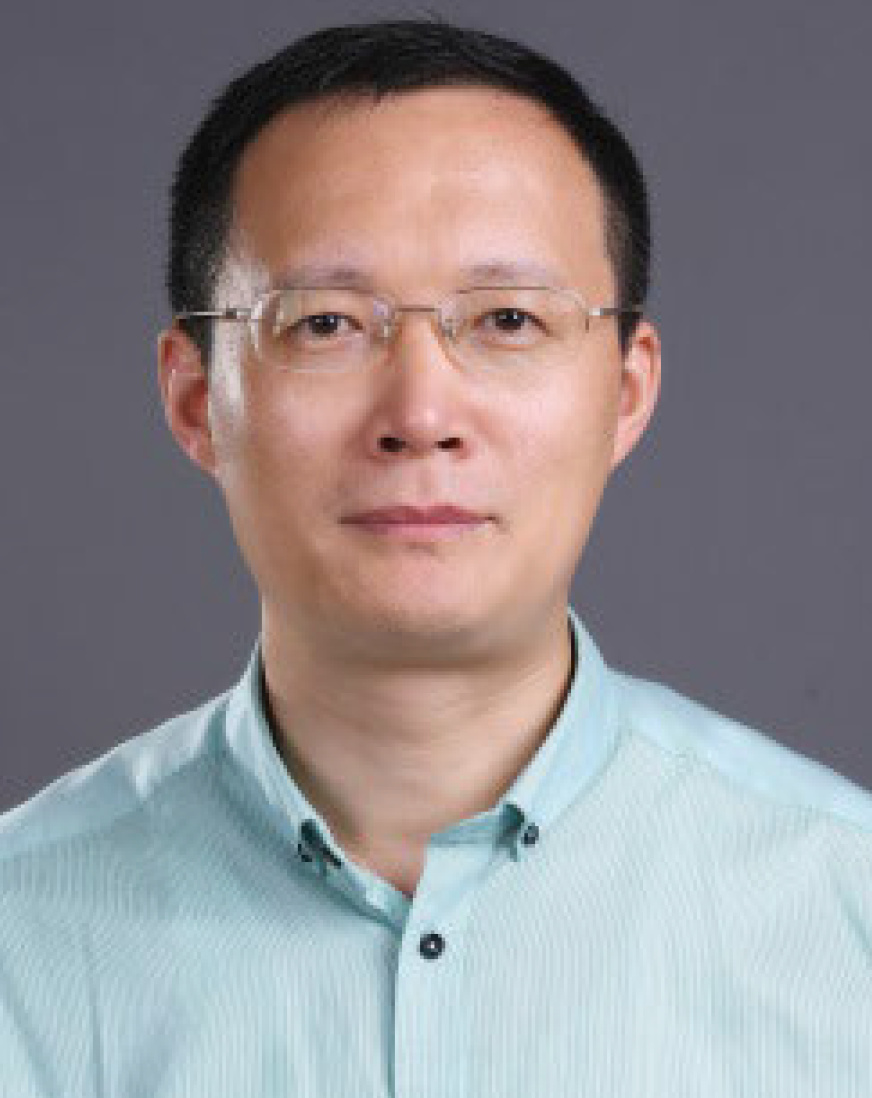}}]{Liqiang Zhang} (Member, IEEE) received the Ph.D. degree in geoinformatics from Chinese Academy of Science’s Institute of Remote Sensing Applications, Beijing, China, in 2004. \\
    He is currently a Professor at the School of Geography, Beijing Normal University, Beijing. His research interests include remote sensing image processing, 3-D urban reconstruction, and spatial object recognition.
    \end{IEEEbiography}
    \begin{IEEEbiography}[{\includegraphics[width=1in, height=1.25in, clip, keepaspectratio]{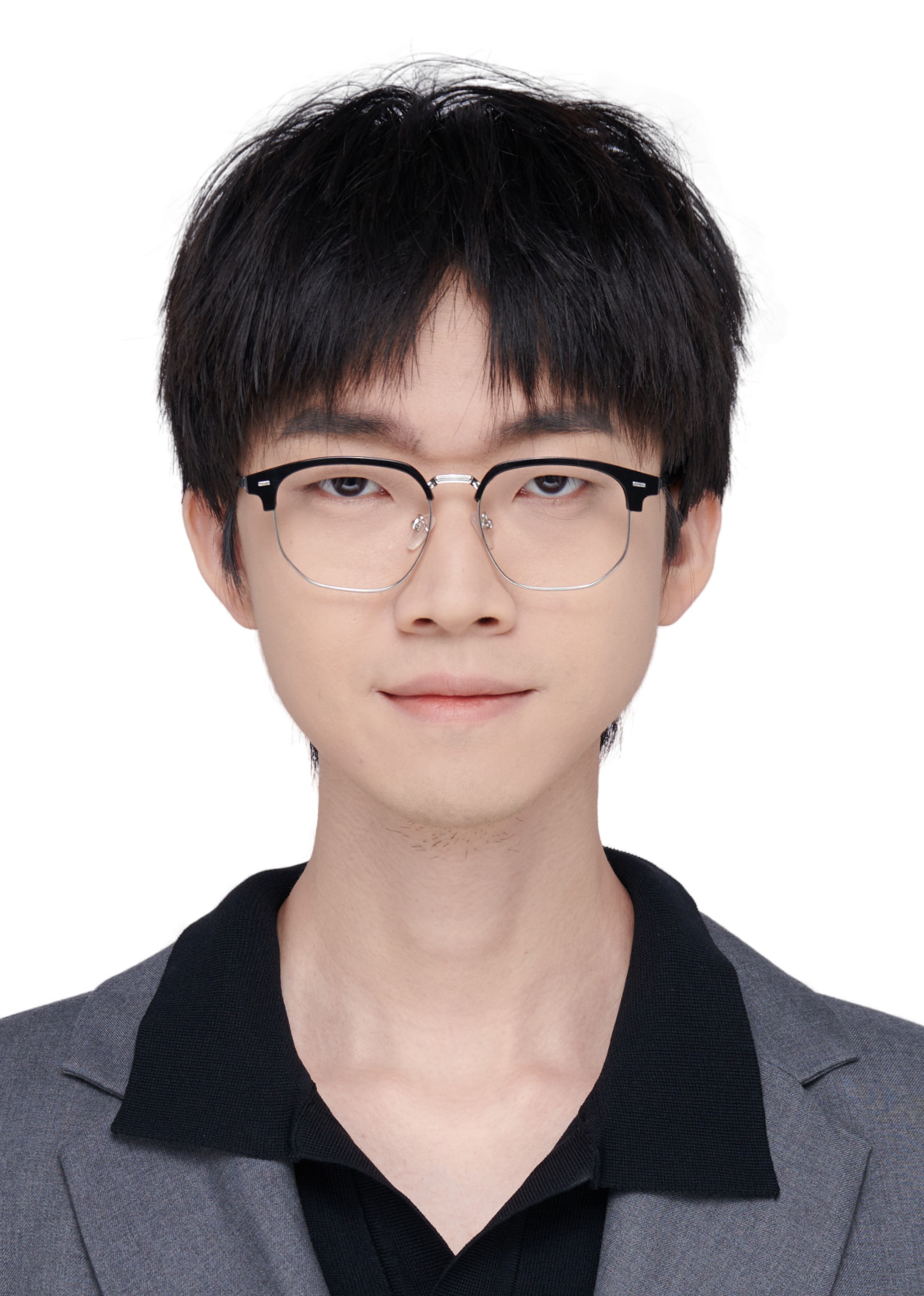}}]{Zhengyang Hou}
        is currently pursuing a Ph.D. in Cartography and Geographic Information Systems at the Faculty of Geographical Science, Beijing Normal University. His research focuses on building damage detection, El Niño and wetland changes, and global land surface albedo variations.
    \end{IEEEbiography}
    \begin{IEEEbiography}[{\includegraphics[width=1in, height=1.25in, clip, keepaspectratio]{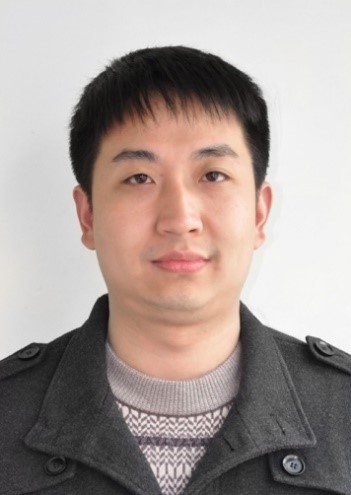}}]{Jun Liu}
        received the B.S. degree in mathematics from the Hunan Normal University, China in 2004. He received the M.S. and Ph.D. degrees in computational mathematics from the Beijing Normal University (BNU), China, in 2008 and 2011 respectively. He is currently an associate professor at BNU. His research interests include variational, optimal transport and deep learning based image processing algorithms and their applications.
    \end{IEEEbiography}
\end{document}